\documentclass[conference]{IEEEtran}
\usepackage{mathtools} 
\usepackage{graphicx} 

\newcounter{tisctr}
\renewcommand{\thetisctr}{\arabic{tisctr}}

\DeclarePairedDelimiter{\norm}{\lvert}{\rvert}
\DeclarePairedDelimiter{\normF}{\lVert}{\rVert}

\newcommand{\spara}[1]{\smallskip\noindent\textbf{#1}}

	{\normalsize \end{list} 
		\vskip 1pt}

\newenvironment{defn-eqn}[2]{\vskip 6pt \refstepcounter{tisctr} %
	\noindent {\textbf{Definition \thetisctr.~}(#1) \emph{#2}} \vspace{2mm}}
	
\newenvironment{defn-test}[2]{\vskip 6pt \refstepcounter{tisctr} %
	\noindent {{\bf Definition \thetisctr.~}(#1) \emph{#2}}%
	\vskip 0pt
	\begin{equation}{}{\labelwidth 0pt \labelsep 0pt%
			\parsep 0pt
		}}%
		{\normalsize \end{equation} 
	    \vspace{-1mm}
		}	
		
\newenvironment{lemma}[1]{\vskip 10pt \refstepcounter{tisctr} %
	\noindent {\textbf{Lemma \thetisctr.~} \emph{#1}}
	\vskip 8pt}	
	
\newcommand{\squishlist}{
	\begin{list}{$\bullet$}
		{ \setlength{\itemsep}{0pt}
			\setlength{\parsep}{1pt}
			\setlength{\topsep}{1pt}
			\setlength{\partopsep}{0pt}
			\setlength{\leftmargin}{1.5em}
			\setlength{\labelwidth}{1em}
			\setlength{\labelsep}{0.5em} } }
	
	\newcommand{\squishend}{\end{list}}

\usepackage{amsmath}
\usepackage{amssymb}
\usepackage{xstring}
\usepackage{booktabs}
\usepackage{adjustbox}
\usepackage{subcaption}
\usepackage{colortbl}
\usepackage{mathtools}
\usepackage{url}
\usepackage[square,comma,numbers]{natbib}
\graphicspath{{./plots/}}
\usepackage{balance}

\newcommand{\comment}[1]{}

\begin{document}

\title{iFair: Learning Individually Fair Data Representations for Algorithmic Decision Making}

\author{\IEEEauthorblockN{Preethi Lahoti}
	\IEEEauthorblockA{Max Planck Institute for Informatics \\
		Saarland Informatics Campus\\
		Saarbr\"{u}cken, Germany\\
		plahoti@mpi-inf.mpg.de}	
	\and
	\IEEEauthorblockN{Krishna P. Gummadi}
	\IEEEauthorblockA{Max Planck Institute for Software Systems\\
		Saarland Informatics Campus\\
		Saarbr\"{u}cken, Germany\\
		gummadi@mpi-sws.org}
	\and
	\IEEEauthorblockN{Gerhard Weikum}
	\IEEEauthorblockA{Max Planck Institute for Informatics\\
		Saarland Informatics Campus\\
		Saarbr\"{u}cken, Germany\\
		weikum@mpi-inf.mpg.de}
}

%


\maketitle

\begin{abstract}
	People are rated and ranked, towards algorithmic decision making
	in an increasing
	number of applications,
	typically based on
	machine learning.
	Research on how to incorporate fairness into
	such tasks has prevalently pursued the paradigm of {group fairness}:
	giving adequate success rates to specifically
	protected groups.
	In contrast, the alternative paradigm of {\em individual fairness} has received
	relatively little attention, 
	and this paper 
	advances this less explored direction.
	The paper introduces a method for probabilistically mapping user records
	into a low-rank representation that 
	reconciles individual fairness and 
	the utility of classifiers and rankings in
	downstream applications.
	Our notion of individual fairness requires that users who are similar
	in all task-relevant attributes such as job qualification, and disregarding
	all potentially discriminating attributes such as gender,
	should have similar outcomes. 	
	We demonstrate the 
	versatility of our method by 
	applying it to
	classification and learning-to-rank tasks on 
	a variety of real-world datasets. 
	Our experiments show substantial improvements over the best prior work
	for this setting.
	
	\noindent
	\textcolor{red}{This is a preprint of a full paper at ICDE 2019. Please cite the ICDE proceedings version.}
	
\end{abstract}
\section{Introduction}

\noindent {\bf Motivation:}
People are rated, ranked and selected or not selected in an increasing number of online
applications, towards algorithmic decisions based on machine learning models.
Examples are approvals or denials of loans or visas, predicting recidivism for law enforcement,
or rankings in job portals.
As algorithmic decision making becomes pervasive in all aspects of our daily life, societal and ethical 
concerns 
\cite{angwin2016machine,crawford2016there} 
are rapidly growing.
A basic approach is to establish policies that disallow the inclusion of potentially
discriminating attributes such as gender or race,
and
ensure that classifiers and rankings operate solely on task-relevant attributes such as
job qualifications.



The problem has garnered significant attention in the data-mining and machine-learning communities. 
Most of this work considers so-called {\em group fairness} models, 
most notably, the statistical parity of outcomes in binary classification tasks, as a notion of fairness.
Typically, classifiers are extended to incorporate demographic groups in their loss functions,
or include constraints on the fractions of groups in the accepted class
\cite{calders2009building,kamiran2010discrimination,
kamishima2012considerations,pedreshi2008discrimination,
feldman2015certifying,fish2016confidence}
to reflect legal boundary conditions and regulatory policies.
For example, computing a shortlist of people invited for job interviews should have a gender mix
that is proportional to the base population of job applicants.

The classifier objective is faced with a fundamental trade-off between utility
(typically accuracy) and fairness, and needs to aim for a good compromise.
Other definitions of group fairness 
have been proposed \cite{hardt2016equality,zafar2017fairness,zhang2016identifying},
and variants of group fairness have been applied to learning-to-rank tasks \cite{zehlike2017fa,yang2017measuring,singh2018fairness}.
In all these cases, fair classifiers or regression models need an explicit specification of
sensitive attributes such as gender, and often the identification of a specific {\em protected
(attribute-value) group} such as gender equals female.

\noindent {\bf The Case for Individual Fairness:}
\citet{dwork2012fairness} argued that group fairness, while appropriate
for policies regarding demographic groups, does not capture the
goal of treating individual people in a fair manner.
This led to the definition of {\em individual fairness}: 
similar individuals should be treated similarly. 
For binary classifiers,
this means that individuals who are similar on the task-relevant attributes
(e.g., job qualifications) should have nearly the same probability of being
accepted by the classifier.
This kind of fairness is intuitive and captures
aspects that group fairness 
does not handle.
Most importantly, it addresses potential discrimination of people by disparate
treatment despite the same or similar qualifications
(e.g., for loan requests, visa applications or job offers),
and it can 
mitigate such risks.


\noindent{\bf Problem Statement:}
Unfortunately, the rationale for capturing individual fairness 
has not received much follow-up work --
the most notable exception being \cite{zemel2013learning} 
as discussed below.
The current paper advances the approach of individual fairness
in its practical viability, 
and specifically addresses the key problem of 
coping with the critical trade-off between
fairness and utility:
{\em How can a data-driven system provide a high degree of individual
fairness while also keeping the utility of classifiers and
rankings high?}
Is this possible in an application-agnostic manner, so that
arbitrary downstream applications are supported? 
Can the system handle situations
where sensitive attributes are not explicitly specified at all
or become known
only at decision-making time (i.e., after the system was trained
and deployed)?

Simple approaches like removing all sensitive attributes from the data and
then performing a standard clustering technique do not reconcile
these two conflicting goals, as standard clustering may lose too
much utility and individual fairness needs to consider attribute
correlations beyond merely masking the explicitly protected ones.
Moreover, the additional goal of generality, in terms of supporting
arbitrary downstream
applications, mandates that cases without explicitly sensitive attributes
or with sensitive attributes being known only at decision-making time
be gracefully handled as well.

The following example illustrates the points that 
a) individual fairness addresses situations that group fairness
does not properly handle, and 
b) individual fairness must be carefully traded off against
the utility of classifiers and rankings.


	
\noindent {\bf Example:} 
Table \ref{tbl-xing-ranking-statistical-parity} shows a real-world example for the issue of
unfairness to individual people. 
Consider the ranked results for an employer's
query ``Brand Strategist'' on
the German job portal \emph{Xing}; 
that data was originally used in \citet{zehlike2017fa}.
The top-10 results satisfy 
group fairness with regard to gender, as defined by \citet{zehlike2017fa} where
a top-k ranking $\tau$ is fair if for every prefix $\tau|i = <\tau(1), \tau(2), \cdots \tau(i)>$ 
($1 \leq i \leq k$) the set $\tau|i$ 
satisfies statistical parity with statistical significance $\alpha$. 
However the outcomes 
in Table \ref{tbl-xing-ranking-statistical-parity} are far from being fair for the individual users:  people with very similar qualifications, such as
Work Experience and Education Score ended up on ranks that are far
apart (e.g., ranks 5 and 30). By the \emph{position bias} \cite{JoachimsR07}
when searchers browse result lists,
this treats the low-ranked people quite unfairly.
This demonstrates that applications can satisfy group-fairness policies, 
while still being unfair to individuals.
	
	\begin{table}[t!]
		\centering
		\tiny
		\noindent\resizebox{0.9\columnwidth}{!}{%
			\setlength\tabcolsep{1.5pt} 
	\begin{tabular}{@{}ccccc@{}}
		\toprule
		Search Query                         & Work       & Education & Candidate & Xing    \\
		                                     & Experience & Experience     &           & Ranking \\ \midrule
		Brand Strategist                     & 146        & 57        & male      & 1       \\
		Brand Strategist                     & 327        & 0         & female    & 2       \\
		Brand Strategist                     & 502        & 74        & male      & 3       \\
		Brand Strategist                     & 444        & 56        & female    & 4       \\
		\rowcolor[gray]{.8} Brand Strategist & 139        & 25        & male      & 5       \\
		Brand Strategist                     & 110        & 65        & female    & 6       \\
		Brand Strategist                     & 12         & 73        & male      & 7       \\
		Brand Strategist                     & 99         & 41        & male      & 8       \\
		Brand Strategist                     & 42         & 51        & female    & 9       \\
		Brand Strategist                     & 220        & 102       & female    & 10      \\
		&            & $\cdots$  &           &         \\
		Brand Strategist                     & 3          & 107       & female    & 20      \\
		\rowcolor[gray]{.8} Brand Strategist & 123        & 56        & female    & 30      \\
		Brand Strategist                     & 3          & 3         & male      & 40      \\ \bottomrule
	\end{tabular}
			}
	\caption{Top k results on \protect\url{www.xing.com} (Jan 2017) for an employer's job search query ``Brand Strategist''.}
			\label{tbl-xing-ranking-statistical-parity}			
			\vspace{-7mm}
		\end{table}

\noindent{\bf State of the Art and its Limitations:}
Prior work on fairness for ranking tasks has exclusively focused on group fairness \cite{zehlike2017fa,yang2017measuring,singh2018fairness},
disregarding the dimension of individual fairness.
For the restricted setting of binary classifiers, the most notable work on 
individual fairness is  \cite{zemel2013learning}.
That work addresses the fundamental trade-off between utility and fairness
by defining a combined loss function to learn a low-rank data representation.
The loss function reflects a 
weighed sum of 
classifier accuracy, statistical parity 
for a single pre-specified 
protected group, and individual fairness in terms of 
reconstruction loss of data.
This model, called {\em LFR}, 
is powerful and elegant,  
but has major limitations:
\squishlist
\item It is geared for binary classifiers and does not generalize to
a wider class of machine-learning tasks, 
dismissing regression models, i.e., learning-to-rank tasks.
\item Its data representation is tied to a specific use case with
a single protected group that needs to be specified upfront.
Once learned, the representation cannot be dynamically adjusted to different
settings later.
\item Its objective function strives for a compromise over three components:
application utility (i.e., classifier accuracy), group fairness and individual fairness.
This tends to burden
the learning with too many aspects that cannot be reconciled.
\squishend

%
%
%
%

Our approach
overcomes these
limitations by developing a
model for representation learning that focuses on
individual fairness and offers greater flexibility and versatility.

\noindent {\bf Approach and Contribution:}
%
The approach that we put forward in this paper, called {\em iFair}, is to learn
a generalized data representation that preserves the fairness-aware similarity between individual records
while also aiming to minimize or bound the data loss.
This way, we aim to reconcile individual fairness and application utility,
and we intentionally disregard group fairness as an explicit criterion.

iFair 
resembles the model of \cite{zemel2013learning} in that we also learn a 
representation 
via
probabilistic clustering, using a form of gradient descent
for optimization.
However, our approach differs from \cite{zemel2013learning} 
on a number of major aspects:
\squishlist
\item iFair learns flexible and versatile
representations, instead of committing to a specific downstream application
like binary classifiers.
This way, we open up applicability to arbitrary classifiers and support
regression tasks (e.g., rating and ranking people) as well.
\item iFair does not depend on a pre-specified protected group.
Instead, it supports multiple sensitive attributes where the 
``protected values'' are known only at run-time
after the application is deployed. 
For example, we can easily handle situations where the critical value
for gender is female for some ranking queries and male for others.
%
\item iFair does not consider any notion of group fairness in its
objective function. 
This design choice relaxes the optimization problem, and we achieve
much better utility with very good fairness in both classification and ranking tasks.
Hard group-fairness constraints, based on legal requirements,
can be enforced post-hoc by adjusting the outputs of 
iFair-based classifiers or rankings. 
\squishend

\begin{figure*}[thb!]
	\centering	
	\includegraphics[scale=0.35]{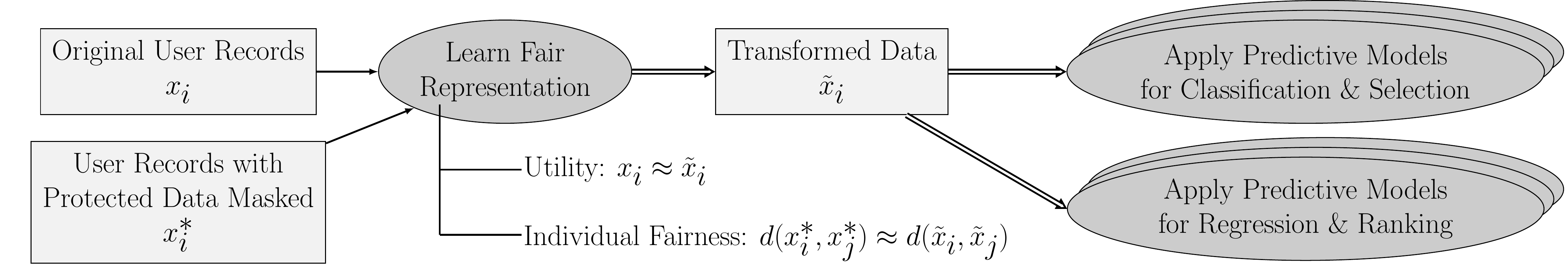}
	\caption{Overview of decision-making pipeline.} 
	\label{fig:flowchart}
	\vspace{-2mm}
\end{figure*}

The novel contributions of iFair are:
1) the first method, to the best of our knowledge, that provides
individual fairness for learning-to-rank tasks;
2) an application-agnostic framework for learning low-rank data representations
that reconcile individual fairness and utility such that
application-specific choices on sensitive attributes and values 
do not require learning another representation;
3) experimental studies with classification and regression tasks for
downstream applications, empirically showing that iFair
can indeed reconcile strong individual fairness with high utility. 
The overall decision-making pipeline
is illustrated in Figure \ref{fig:flowchart}.

\section{Related work}\label{section:related-work}
\spara{Fairness Definitions and Measures:} Much of the work in algorithmic fairness has focused on supervised machine learning, specifically on the case of binary classification tasks. 
Several notions of \emph{group fairness} have been proposed in the literature. The most widely used 
criterion
is 
{\em statistical parity} and its 
variants \cite{calders2009building,kamiran2010discrimination,kamishima2012considerations,pedreshi2008discrimination,feldman2015certifying,fish2016confidence}. 
Statistical parity states that the predictions
$\hat{Y}$ of a classifier are 
fair if members of sensitive subgroups, such as people of
certain nationalities or ethnic backgrounds, have an
acceptance likelihood proportional to their share in the
entire data population. 
This is equivalent to requiring that
apriori knowledge of the classification outcome of an individual should provide no information about her membership to 
such subgroups. 
However, for many applications, such as risk assessment
for credit worthiness,
statistical parity is neither feasible nor desirable.

Alternative notions of group fairness have been defined. \citet{hardt2016equality} proposed {\em equal odds} 
which 
requires that the rates of true positives and false positives be the same across groups. 
This punishes classifiers which perform well only on 
specific groups. \citet{hardt2016equality} also proposed a relaxed version of equal odds called 
{\em equal opportunity} which demands only the equality of true positive rates. 
Other definitions of group fairness 
include \emph{calibration} \cite{flores2016false, kleinberg_et_al:LIPIcs:2017:8156}, \emph{disparate mistreatment} \cite{zafar2017fairness}, and \emph{counterfactual fairness} \cite{DBLP:conf/nips/KusnerLRS17}.
Recent work
highlights the inherent incompatibility 
between several notions of group fairness
 and the impossibility of achieving them simultaneously \cite{kleinberg_et_al:LIPIcs:2017:8156,chouldechova2017fair,friedler2016possibility,corbett2017algorithmic}. 


\citet{dwork2012fairness} gave the first 
definition of \emph{individual fairness},
arguing for the fairness of outcomes for individuals and not merely as a group statistic. 
Individual fairness mandates that \emph{similar individuals should be treated similarly}. 
\cite{dwork2012fairness} further 
develops a theoretical framework for mapping individuals to a probability distribution over outcomes, which satisfies 
the Lipschitz property (i.e., distance preservation)
in the mapping. In this paper, we follow up on this definition of individual fairness and present a generalized framework for learning individually fair representations of the data.


\spara{Fairness in Machine Learning:} A parallel line of work in the area of algorithmic fairness uses a specific definition of fairness in order to design fairness models that achieve fair outcomes. 
To this end, there are two general strategies. The first strategy consists of {\em de-biasing} the input data by 
appropriate preprocessing
\cite{kamiran2010discrimination,pedreshi2008discrimination,feldman2015certifying}. 
This typically involves data perturbation such as modifying the value of sensitive attributes or class labels in the training data to satisfy certain fairness conditions,
 such as equal proportion of positive (negative) class labels in both protected and non-protected groups .
The second strategy consists of designing fair algorithmic models - based on \emph{constrained optimization}
\cite{calders2009building,kamishima2012considerations,hardt2016equality,zafar2017fairness}. 
Here, fairness constraints are usually introduced as regularization terms in the objective function. 

\spara{Fairness in IR:} Recently, definitions of group fairness have been extended to learning-to-rank tasks. \citet{yang2017measuring} introduced statistical parity in rankings. \citet{zehlike2017fa} built on \cite{yang2017measuring} and proposed to ensure 
statistical parity at all top-k prefixes of the ranked results. \citet{singh2018fairness} proposed a generalized fairness framework for a larger class of group fairness definitions (e.g., disparate treatment and disparate impact).
However, all this prior work has focused on group fairness alone.
It implicitly assumes that individual fairness is taken care
of by the ranking quality, disregarding situations where
trade-offs arise between these two dimensions.
The recent work of \citet{Biega:SIGIR2018} 
addresses individual fairness in rankings from the
perspective of giving fair exposure to items over
a series of rankings, thus mitigating the position bias
in click probabilities.
In their approach they explicitly assume to have access to scores that are already  individually fair. As such, their work is complementary to ours as they do not address how such a score, which is individually fair can be computed.

\spara{Representation Learning:} 
The work of \citet{zemel2013learning} is the closest to ours in that  it
is also learns low-rank representations by 
probabilistic mapping of data records. 
However, the methods deviates from our in important ways. First, its fair representations are tied to a particular classifier
by assuming a binary classification problem 
with pre-specified labeling target attribute
and a single protected group.
In contrast, the representations learned by {\em iFair}
are agnostic to the downstream learning tasks
and thus easily deployable for new applications.
Second, the optimization  in \cite{zemel2013learning}
aims to combine
three competing objectives:
classifier accuracy, statistical parity, and data loss
(as a proxy for individual fairness). 
The {\em iFair} approach, on the other hand, 
addresses a more 
streamlined 
objective function
by focusing on classifier accuracy and individual fairness.

Approaches similar to \cite{zemel2013learning} have been applied to learn censored representations for fair classifiers via adversarial training \cite{louizos2015variational,edwards2015censoring}. 
Group fairness criteria
are optimized in the presence of an adversary. 
These approaches do not consider individual fairness at all.
%

\section{Model} \label{section:methodology}
We consider user records that are fed into a 
learning algorithm 
towards algorithm decision making.
A fair algorithm should make its decisions solely based on
non-sensitive attributes (e.g., technical qualification or education) 
and should disregard sensitive attributes
that bear the risk of discriminating users (e.g., ethnicity/race).
This dichotomy of attributes is specified upfront, by domain 
experts and follows legal regulations and policies.
Ideally, one should consider also strong correlations (e.g., geo-area
correlated with ethnicity/race), but this is usually beyond the scope
of the specification.
We start with introducing preliminary notations and definitions.

\spara{Input Data:}
The input data for $M$ users with $N$ attributes is an $M \times N$ matrix
$X$ with binary or numerical values (i.e., after unfolding or encoding
categorical attributes). Without loss of generality, we assume that
the attributes $1~..~l$ are non-protected and
the attributes $l+1~..~N$ are protected.
We denote the $i$-th user record consisting of all attributes as $x_i$ 
and only non-protected attributes as $x_{i}^{\ast}$.
%
Note that, unlike in prior works,
the set of protected attributes is allowed to be empty
(i.e., $l=N$).
Also, we do not assume any upfront specification
of which attribute values form a protected group.
So a downstream application can flexibly decide
on the critical values (e.g., male vs. female or
certain choices of citizenships) on a case-by-case basis.

\spara{Output Data:}
The goal is to transform the input records  $x_i$
into representations $\tilde{x_i}$ that are directly usable by
downstream applications and have better properties regarding fairness.
Analogously to the input data, we can write the entire output of $\tilde{x_i}$ 
records as an $M \times N$ matrix $\tilde{X}$.

\spara{Individually Fair Representation:} Inspired by the \citet{dwork2012fairness} notion of individual fairness, \textit{``individuals who are similar 
should be treated similarly''}, we propose the following definition for individual fairness:

\begin{defn-eqn}{Individual Fairness}{Given a distance function $d$ in the $N-$dimensional data space, a mapping 
$\phi$ of input records $x_i$ into output records $\tilde{x_i}$ is
 individually fair if for every pair $x_i,x_j$ we have
 \smallskip
\begin{equation}\label{defn-individualFairness}
	\norm{d(\phi(x_i),\phi(x_j)) - d(x_{i}^*,x_{j}^*)} \leq \epsilon 
\end{equation}}
\end{defn-eqn}

The definition requires that individuals who are (nearly) indistinguishable
on their non-sensitive attributes in $X$ should also be (nearly) indistinguishable
in their transformed representations $\tilde{X}$. 
For example, two people with (almost) the same technical qualifications for a 
certain job should have (almost) the same low-rank representation, regardless of
whether they differ on protected attributes such as gender, religion or ethnic group.
In more technical terms, a distance
measure between user records should be preserved in the transformed space.

Note that this definition intentionally deviates from the 
original definition of \emph{individual fairness} of \cite{dwork2012fairness} in that with $x_i^*, x_j^*$ 
we consider only the non-protected attributes of the original user records,
as protected attributes should not play a role in the decision outcomes of an individual. 

%
%



\subsection{Problem Formulation: Probabilistic Clustering}

As individual fairness needs to preserve similarities between records $x_i, x_j$,
we cast the goal of computing good
representations $\tilde{x_i},\tilde{x_j}$ into
a formal problem of {\em probabilistic clustering}.
We aim for $K$ clusters, each given in the form of a {\em prototype vector} $v_k$ ($k=1..K$),
such that records $x_i$ are assigned to clusters by a record-specific probability distribution
that reflects the distances of records from prototypes.
This can be viewed as a low-rank representation of the input matrix $X$ with $K < M$,
so that we reduce attribute values into a more compact form.
As always with soft clustering, $K$ is a hyper-parameter.

\begin{defn-eqn}{Transformed Representation}{
The fair representation $\tilde{X}$, an
$M \times N$ matrix of row-vise output vectors $\tilde{x_i}$, consists of
\squishlist
\item[(i)] $K < M$ prototype vectors $v_k$, each of dimensionality $N$, 
\item[(ii)] a probability distribution $u_i$, of dimensionality $K$, for each input record $x_i$ where 
$u_{ik}$ is the probability of $x_i$ belonging to the cluster of prototype $v_k$.
\squishend
The representation $\tilde{x_i}$ is given by
\vspace{2mm}
\smallskip
\begin{equation}\label{eq:transformed-representation}
\tilde{x_i} = \sum_{k=1..K} u_{ik} \cdot v_k
\vspace{2mm}
\end{equation}
or equivalently in matrix form: 
$\tilde{X} = U \times V^T$
where the rows of $U$ are
the per-record probability distributions and the columns of $V^T$ are the
prototype vectors.
}
\end{defn-eqn}
\label{defn:transformed-representation}

\begin{defn-eqn}{Transformation Mapping}{
We denote the
mapping $x_i \rightarrow \tilde{x_i}$ as $\phi$;  that is,
\smallskip
\begin{equation}\label{eq:transformed-mapping}
\phi(x_i) = \tilde{x_i} = \sum_{k=1..K} u_{ik} \cdot v_k
\end{equation}
using 
Equation \ref{eq:transformed-representation}.
}
\end{defn-eqn}

\comment{

We avoid making assumptions on the learning algorithm and downstream
application, striving for a generic and versatile model.
We aim to learn a low-rank representation 
with individual fairness as an optimization goal
while, at the same time, keeping utility as high as possible.

\begin{defn-eqn}{Low-Rank Representation}{A low-rank representation of input data $X$ is a factorization given by
\begin{equation}
	X \approx U V^T =: \tilde{X}
\end{equation}}
\end{defn-eqn}

where 
$U$ is an $M \times K$ matrix and $V^T$ is a $K \times N$ matrix, both of rank $K$ with
$K < N$ and $K < M$.
The $K$ dimensions of the low-rank representation are latent,
and the columns of the resulting $\tilde{X}$ may not be directly interpretable. The column vectors $v_k \in V$ act as prototype vectors in the same $N-$ dimensional space as $x$. Each entry $U_{m,k}$ in the matrix $U$ gives the probability that datapoint $x_m$ maps to 
the $k$-th prototype $v_k$.  Thus $U$ is a $M \times K$ matrix consisting of mixtures of these probabilities. 


}


\spara{\emph{Utility Objective:}}
Without making any assumptions on the downstream application,
the best way of ensuring high utility is to minimize the data loss
induced by $\phi$.
\begin{defn-eqn}{Data Loss}{The reconstruction loss between $X$ and $\tilde{X}$ 
is
the sum of squared errors
\smallskip
\begin{equation}
	L_{util}(X,\tilde{X}) = \sum\limits_{i = 1}^{M} ||x_{i} - \tilde{x}_{i}||_2 = 
\sum\limits_{i = 1}^{M} \sum\limits_{j = 1}^{N} (x_{ij} - \tilde{x}_{ij})^2
\end{equation}}	
\end{defn-eqn}

\spara{\emph{Individual Fairness Objective:}}
Following the rationale for Definition \ref{defn-individualFairness},
the desired transformation $\phi$
should preserve pair-wise distances between data records on
non-protected attributes. 

\begin{defn-eqn}{Fairness Loss}{ For input data $X$, with row-wise data records $x_i$, 
and its transformed representation $\tilde{X}$ with row-wise $\tilde{x_i}$,
		the fairness loss  $L_{fair}$ is
\smallskip
\begin{equation}
	L_{fair}(X,\tilde{X})  = \sum_{i,j=1..M} \left(d(\tilde{x_i},\tilde{x_j}) - d(x^*_i,x^*_j)\right)^2
\end{equation}}
\end{defn-eqn}




\spara{\emph{Overall Objective Function:}}
Combing the data loss and the fairness loss yields our final objective function
that the learned representain should aim to minimize.

\begin{defn-eqn}{Objective Function}{
The combined objective function is given by
\begin{equation}\label{eq:objective}
L = \lambda \cdot L_{util}(X,\tilde{X}) + \mu \cdot L_{fair}(X,\tilde{X})
\end{equation}
where 
$\lambda$ and $\mu$ are hyper-parameters.
}
\end{defn-eqn}\label{defn:objective-function}


\subsection{Probabilistic Prototype Learning}

So far we have left the choice of the distance function
$d$ open. Our methodology is general
and can incorporate a wide suite of distance measures.
However, for the actual optimization, we need to
make a specific choice for $d$.
In this paper, we focus on the family of
{\em Minkowski p-metrics}, which is indeed a metric
for $p \geq 1$. A common choice is $p=2$,  which corresponds 
to a Gaussian kernel.

\begin{defn-eqn}{Distance Function}{
The distance between two data records $x_i,x_j$ is
\begin{equation}\label{eq:distance-function}
d(x_i,x_j) = \big[\sum\limits_{n = 1}^{N} \alpha_n (x_{i,n} - x_{j,n})^p\big]^{1/p}
\end{equation}
where $\alpha$ is an $N$-dimensional vector of
tunable or learnable weights for the different data attributes.
}
\end{defn-eqn}


%
%

This distance function $d$ is applicable to original
data records $x_i$, transformed vectors $\tilde{x_i}$ 
and prototype vectors $v_k$ alike.
In our model, we avoid the quadratic number of
comparisons for all record pairs, and instead 
consider distances only between records and prototype vectors
(cf. also \cite{zemel2013learning}).
Then, these distances can be used to define the
probability vectors $u_i$ 
that hold the probabilities for record $x_i$ belonging to
the cluster with prototype $v_k$ (for $k=1..K$).
To this end, we apply a softmax function to the
distances between record and prototype vectors.

\begin{defn-eqn}{Probability Vector}{
The probability vector $u_i$ for record $x_i$ is
\begin{equation}\label{eq-U_m_k}
u_{i,k} = 
\frac{exp(-d(x_i,v_k))}{\sum\limits_{j=1}^K exp(-d(x_i,v_j))}
\end{equation}
}
\end{defn-eqn}

The mapping $\phi$ that transforms $x_i$ into $\tilde{x_i}$
can be written as
\begin{equation}\label{eq-prototype-mapping} 
\phi(x_i)  =\sum\limits_{k = 1}^{K} 
\frac{exp(-d(x_i,v_k))}{\sum\limits_{j=1}^K exp(-d(x_i,v_j))} \cdot v_k
\end{equation}

With these definitions in place, the task of learning
fair representations $\tilde{x_i}$ now 
amounts to
computing $K$ prototype vectors $v_k$ and 
the $N$-dimensional weight vector $\alpha$ in $d$
such that the overall loss function $L$ 
 is minimized.

\begin{defn-eqn}{Optimization Objective}{
The optimization objective is to compute
$v_k$ ($k=1..K$) and $\alpha_n$ ($n=1..N$)
as argmin for the loss function
\bigskip
\begin{align*}
L& = \lambda \cdot L_{util}(X,\tilde{X}) ~~+~~ \mu \cdot L_{fair}(X,\tilde{X}) \\
& = \lambda \cdot \sum\limits_{i = 1}^{M} \sum\limits_{j = 1}^{N} (x_{ij} - \tilde{x}_{ij})^2 +
\mu \cdot \smashoperator{\sum\limits_{i,j=1..M}} \left(d(\tilde{x_i},\tilde{x_j}) - d(x^*_i,x^*_j)\right)^2 \\
\end{align*}
where $\tilde{x}_{ij}$ and $d$ are substituted using Equations
\ref{eq-prototype-mapping} 
and \ref{eq:distance-function}.
}
\end{defn-eqn}

The $N-$dimensional weight vector $\alpha$ 
controls the influence of each attribute. 
Given our definition of individual fairness (which 
intentionally deviates
from the original definition in \citet{dwork2012fairness}),
a natural setting is to give no weight to the protected attributes
as these should not play any role in the similarity of
(qualifications of) users.
In our experiments, we observe that 
giving (near-)zero weights to the protected attributes 
increases the fairness of the learned data
representations (see Section \ref{section:experiments}).


\subsection{Gradient Descent Optimization:}
Given this setup, the learning system minimizes the 
combined objective function given by
\smallskip
\begin{equation}\label{eq:objective}
L = \lambda \cdot L_{util}(X,\tilde{X}) + \mu \cdot L_{fair}(X,\tilde{X})
\vspace{2mm}
\end{equation} 
where $L_{util}$ is the data loss, $L_{fair}$ is the loss in individual fairness, and $\lambda$ and $\mu$ are hyper-parameters. We have two sets of {\em model parameters} to learn 
\begin{enumerate}
\item[(i)] $v_k$ ($k=1..K$), the $N-$dimensional prototype vectors, 
\item[(ii)] $\alpha$, the $N-$dimensional weight vector of the distance function in Equation  \ref{eq:distance-function}.
\end{enumerate}

\noindent We apply the $L$-$BFGS$ algorithm \cite{liu1989limited}, a quasi-Newton method, to minimize Equation \ref{eq:objective} and learn the model parameters.

\section{Properties of the {\em iFair} Model}
We discuss properties of  {\em iFair} representations
and empirically compare {\em iFair} to the \emph{LFR} model.
We are interested in the general behavior of methods for
learned representations, to what extent they can reconcile
utility and individual fairness at all, and how they relate
to group fairness criteria (although {\em iFair} does not
consider these in its optimization).
To this end, we generate {\em synthetic data} with systematic parameter
variation as follows. 
We restrict ourselves to the case of a 
binary classifier.

\begin{figure*}[tbh!]
	\begin{subfigure}{0.33\linewidth}
		\centering	
		\includegraphics[scale=0.48]{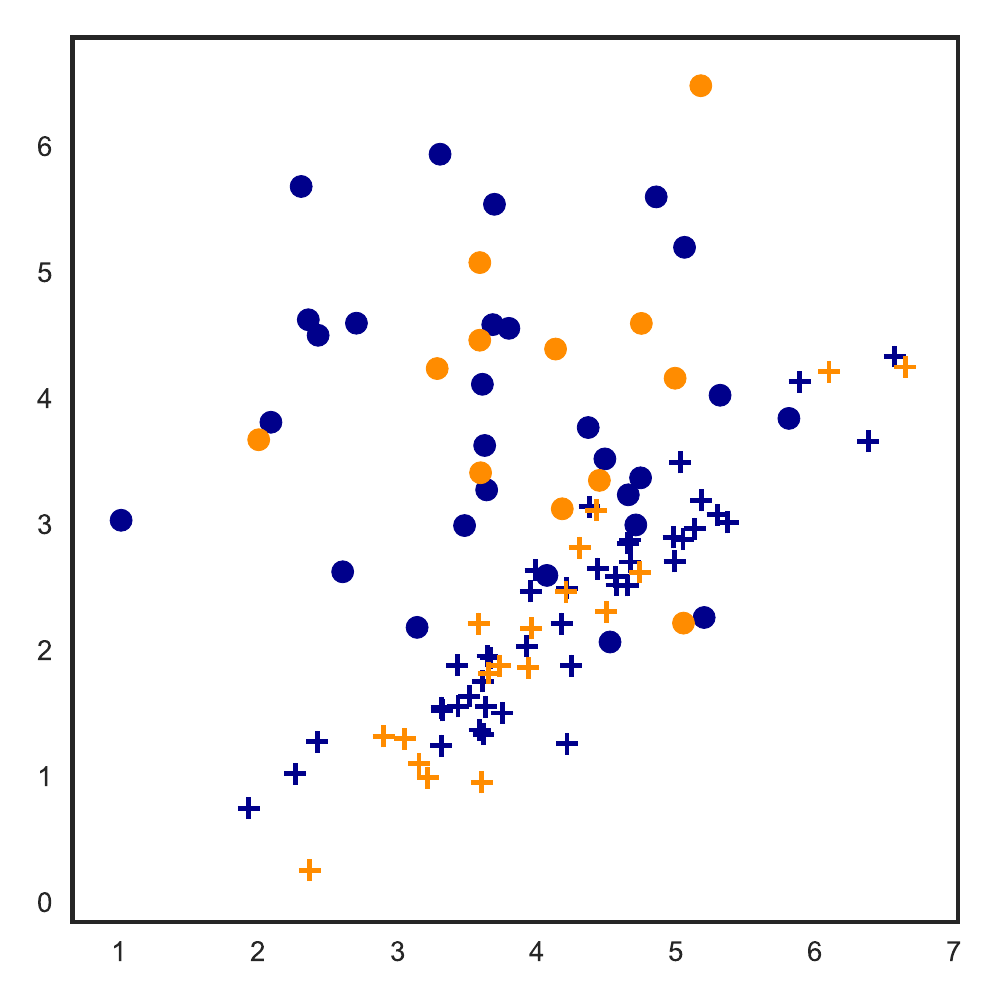}
		\caption{original data (random)}
		\label{fig:synthetic-data-a}
		\vspace{1mm}
	\end{subfigure}	
	\begin{subfigure}{0.33\linewidth}
		\centering	
		\includegraphics[scale=0.48]{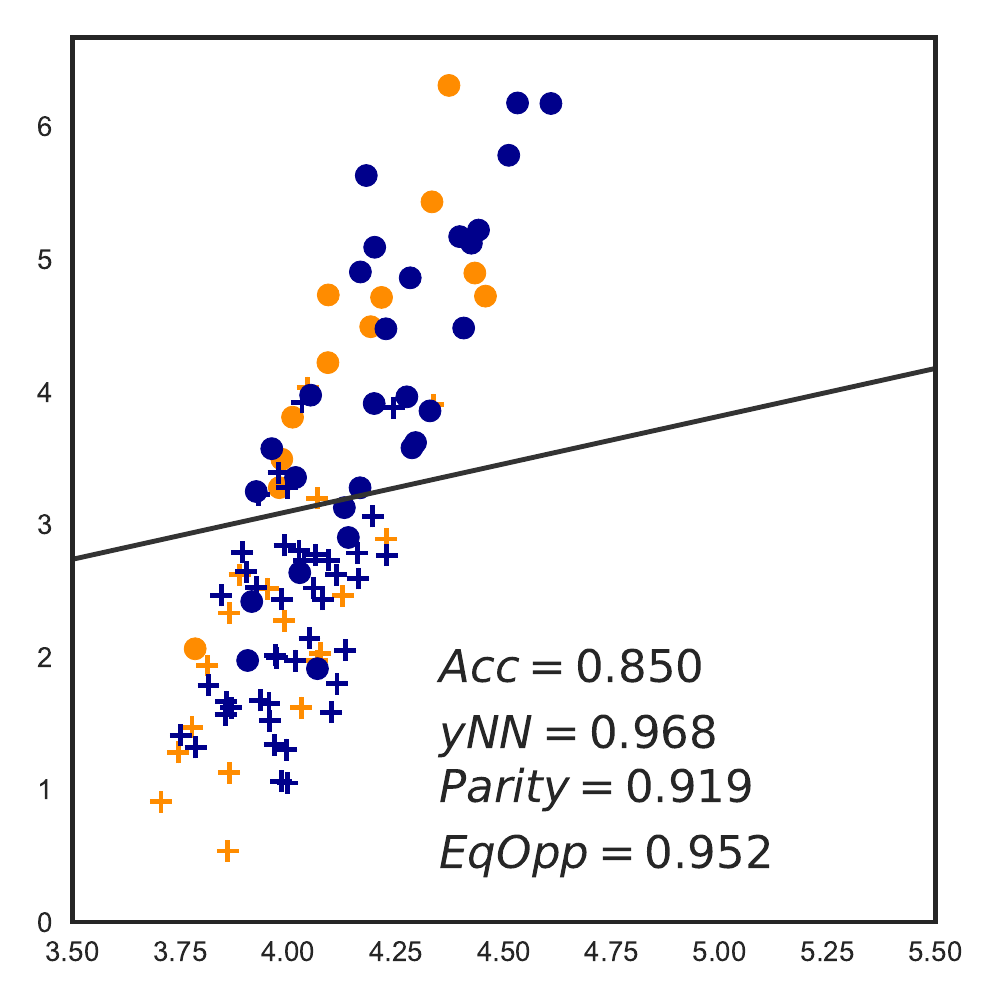}
		\caption{Learned representation via {\em iFair}}
		\label{fig:synthetic-data-b}
		\vspace{1mm}
	\end{subfigure}
	\begin{subfigure}{0.33\linewidth}
		\centering
		\includegraphics[scale=0.48]{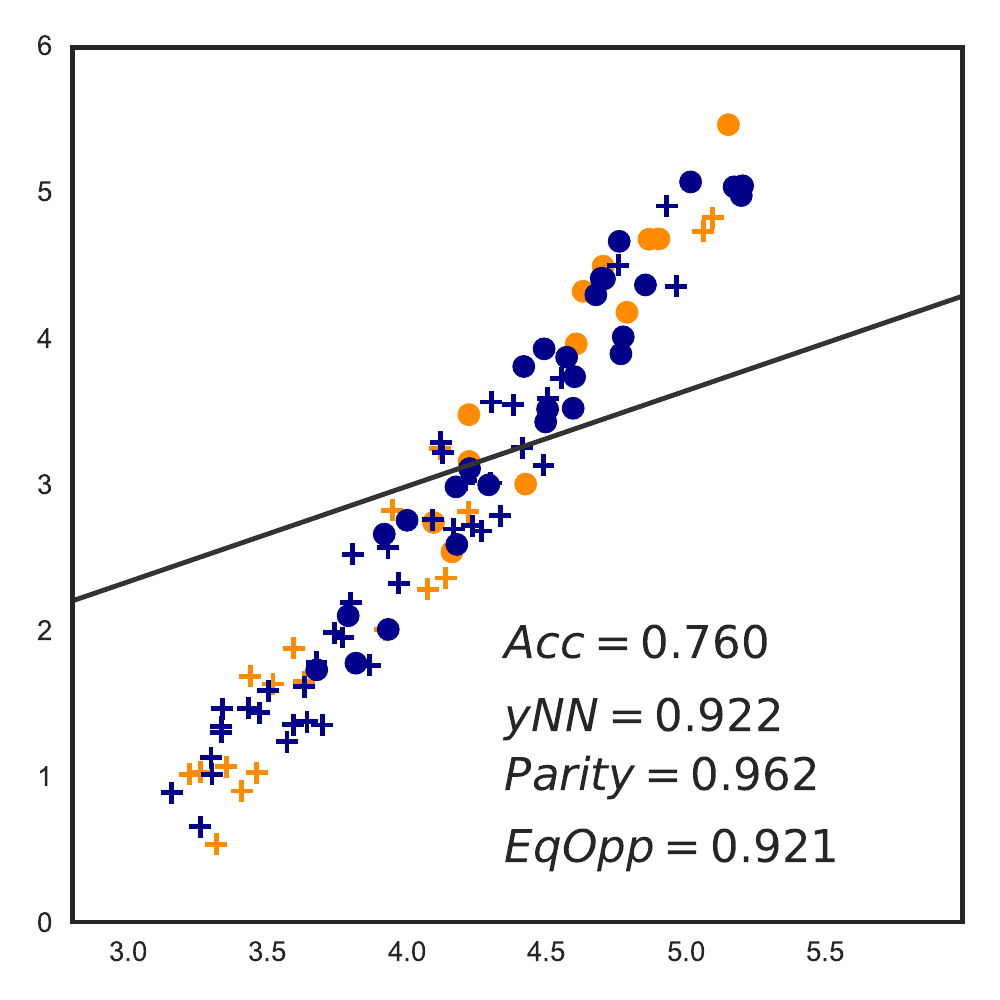}
		\caption{Learned representation via \emph{LFR}}
		\label{fig:synthetic-data-c}
		\vspace{1mm}
	\end{subfigure}
	\newline
	\begin{subfigure}{0.33\linewidth}
		\centering	
		\includegraphics[scale=0.48]{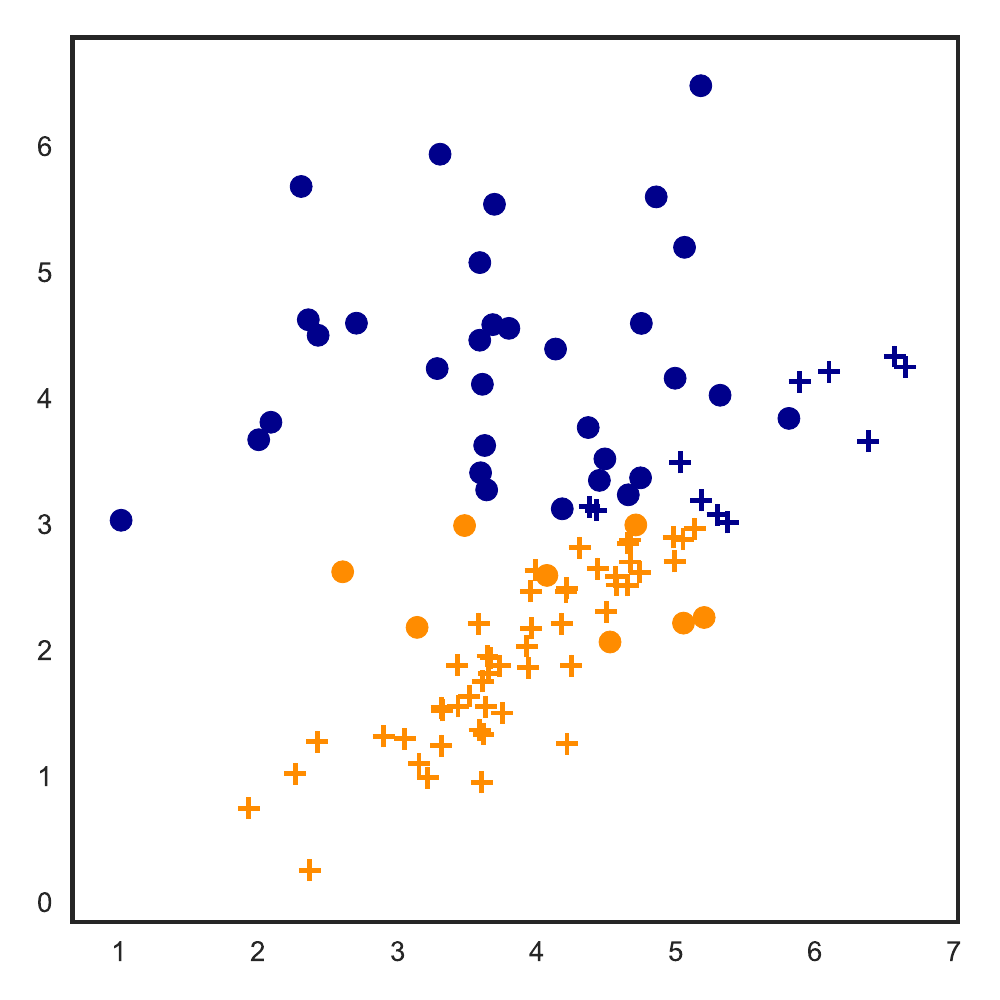}
		\caption{original data ($X1 \leq 3$)}
		\label{fig:synthetic-data-a}
		\vspace{1mm}
	\end{subfigure}	
	\begin{subfigure}{0.33\linewidth}
		\centering	
		\includegraphics[scale=0.48]{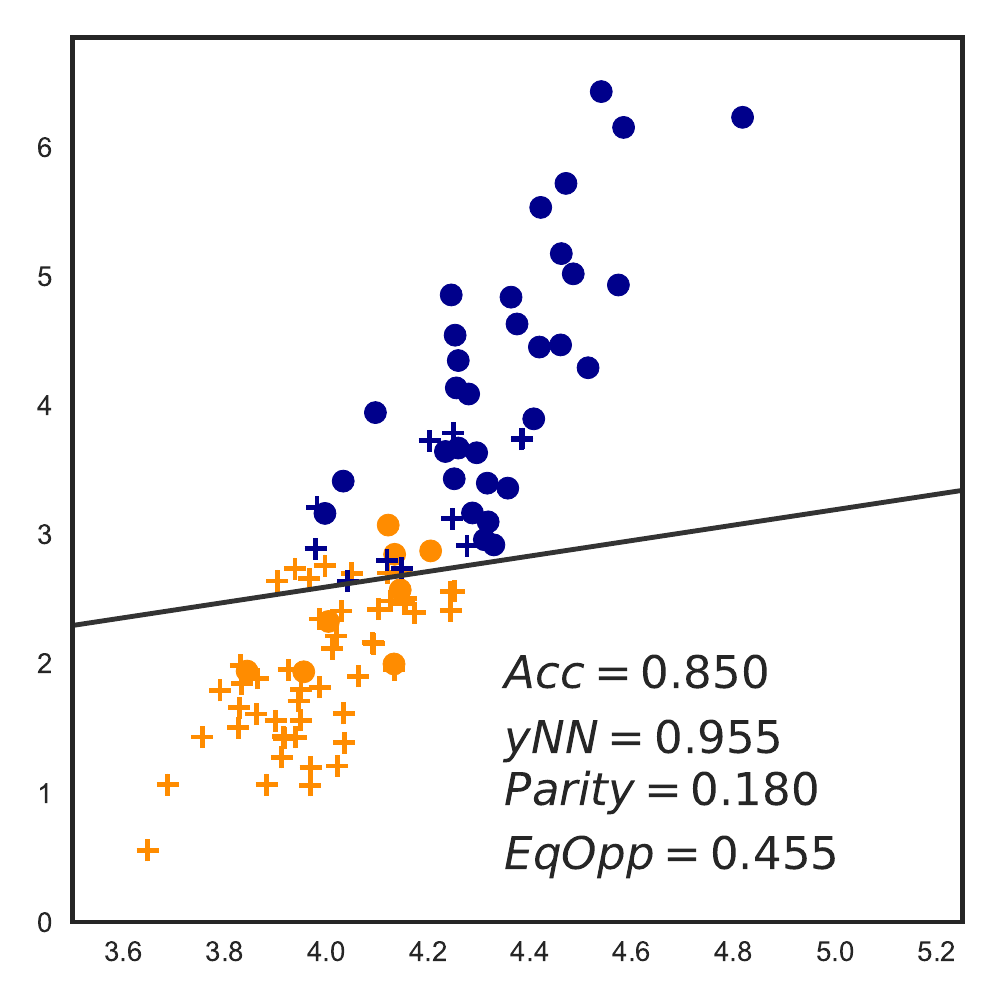}
		\caption{Learned representation via {\em iFair}}
		\label{fig:synthetic-data-b}
		\vspace{1mm}
	\end{subfigure}
	\begin{subfigure}{0.33\linewidth}
		\centering
		\includegraphics[scale=0.48]{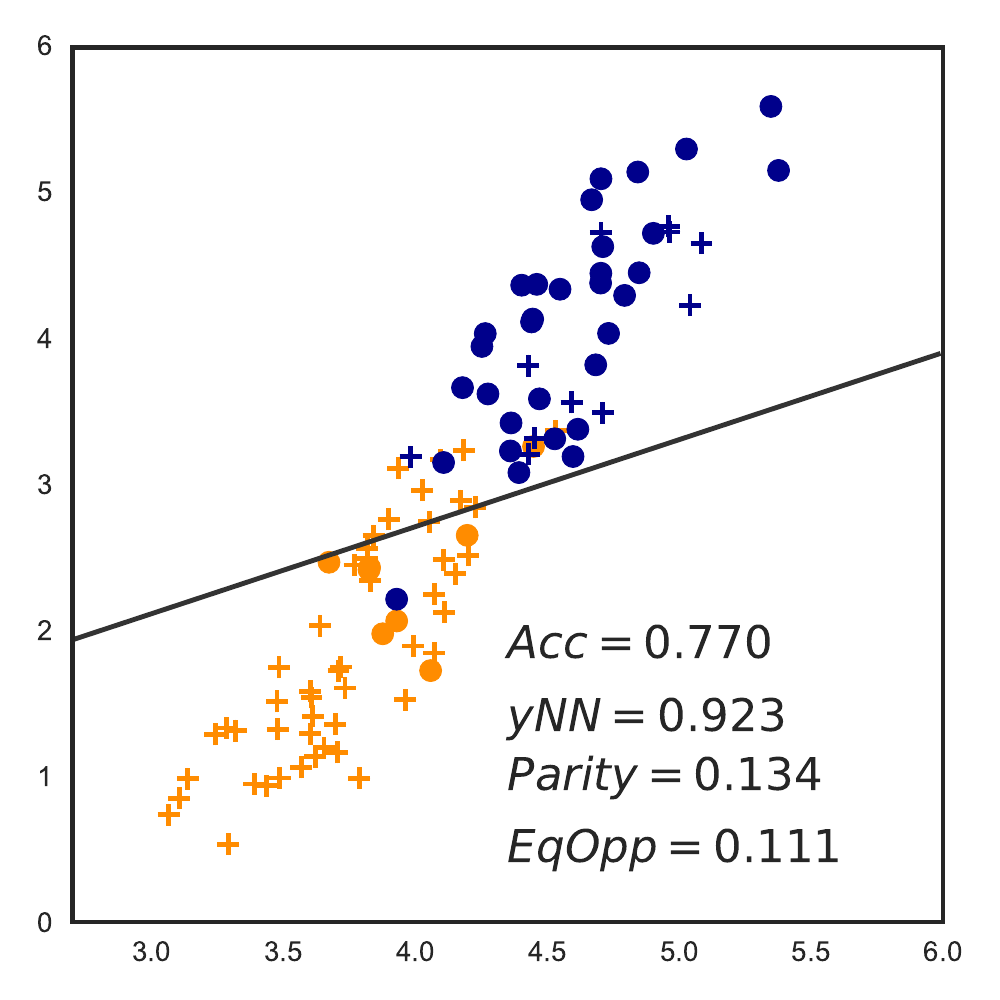}
		\caption{Learned representation via \emph{LFR}}
		\label{fig:synthetic-data-c}
		\vspace{1mm}
	\end{subfigure}
	\newline
	\begin{subfigure}{0.33\linewidth}
		\centering	
		\includegraphics[scale=0.48]{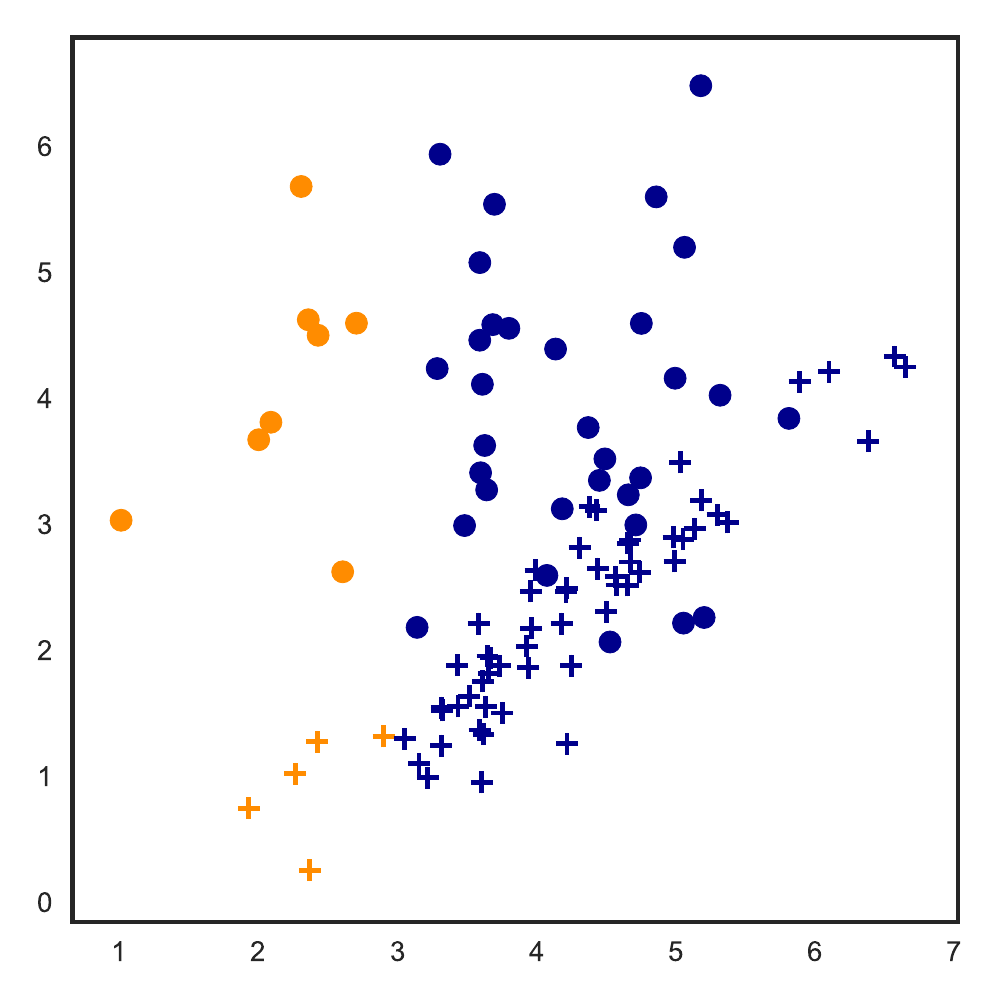}
		\caption{original data ($X2 \leq 3$)}
		\label{fig:synthetic-data-a}
		\vspace{1mm}
	\end{subfigure}	
	\begin{subfigure}{0.33\linewidth}
		\centering	
		\includegraphics[scale=0.48]{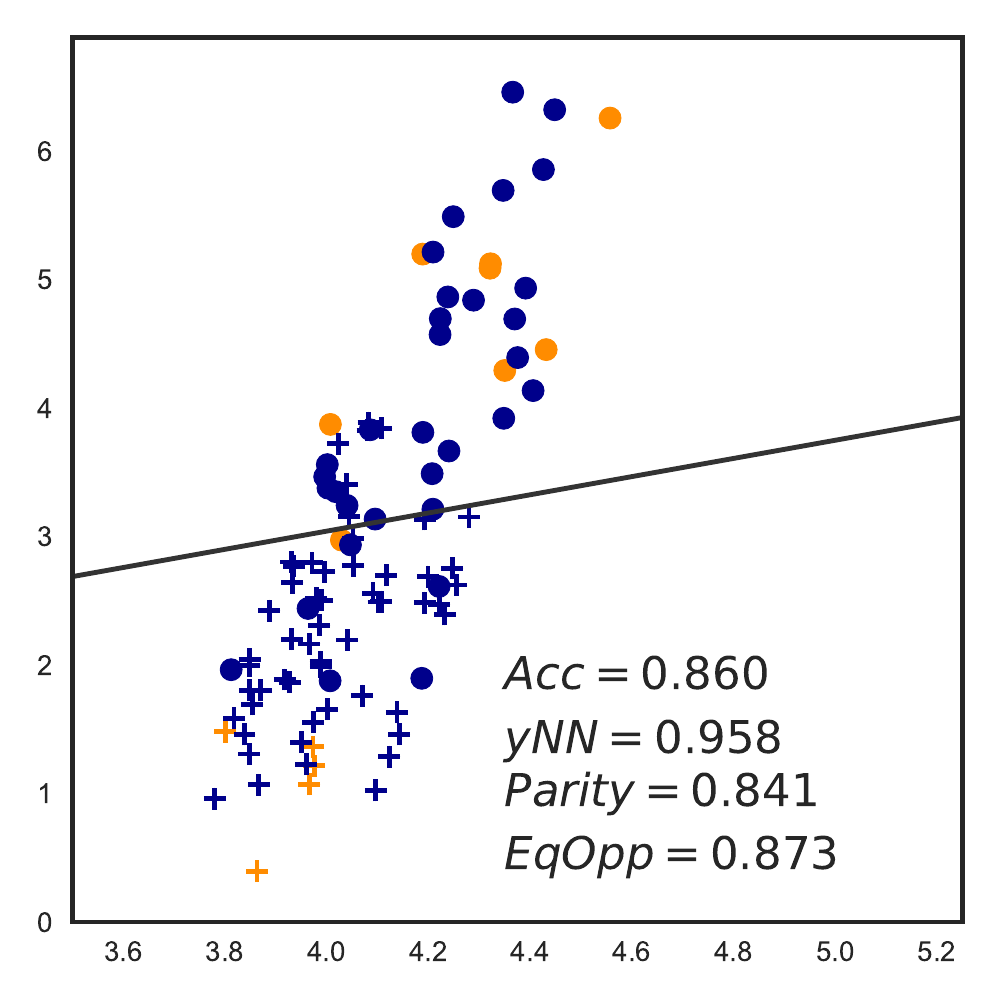}
		\caption{Learned representation via {\em iFair}}
		\label{fig:synthetic-data-b}
		\vspace{1mm}
	\end{subfigure}
	\begin{subfigure}{0.33\linewidth}
		\centering
		\includegraphics[scale=0.48]{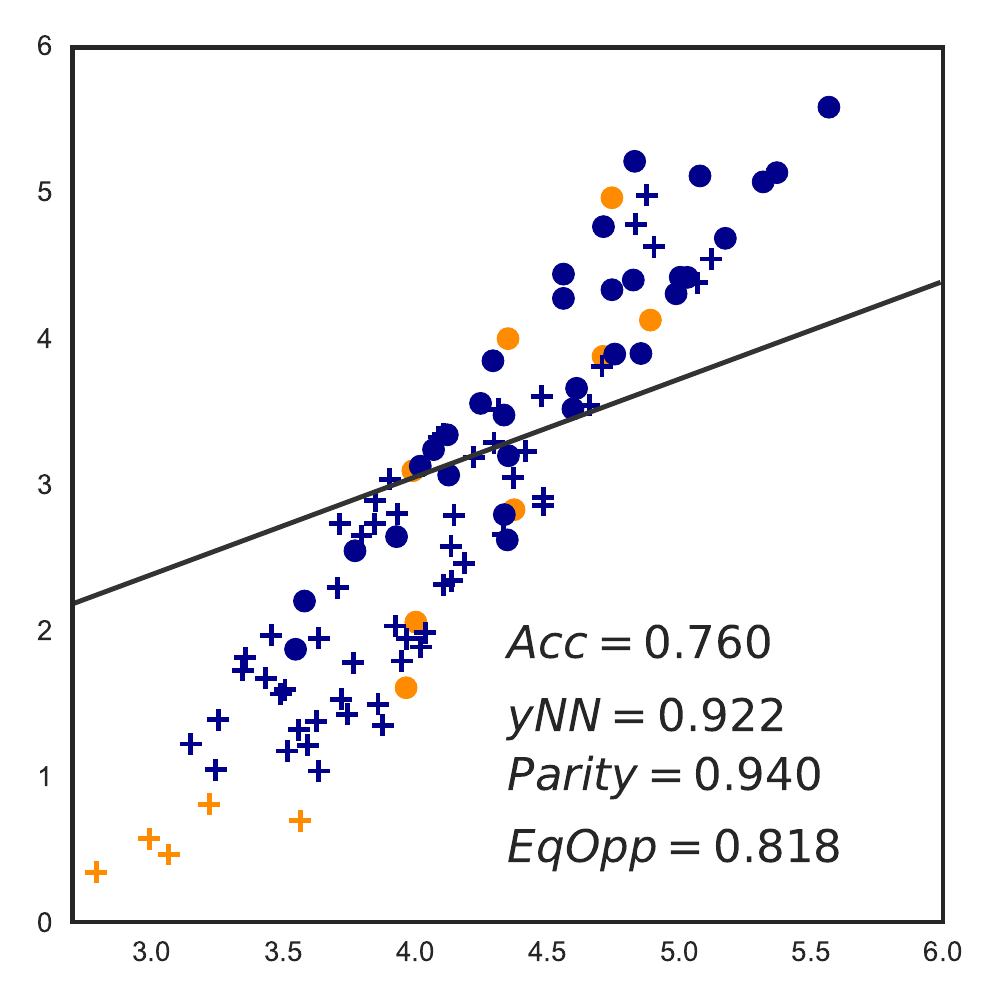}
		\caption{Learned representation via \emph{LFR}}
		\label{fig:synthetic-data-c}
		\vspace{1mm}
	\end{subfigure}
	\caption{Illustration of properties of data representations on synthetic data.
		(left: original data, center: \emph{iFair}, right: \emph{LFR}).
		Output class labels: o for Y=0 and + for Y=1. 
		Membership in protected group: blue for A=0 and orange for A=1.
		Solid lines are the decision boundaries of the respective classifiers. 
		\emph{iFair} outperforms \emph{LFR} on all metrics except for statistical parity.}
	\label{fig:synthetic-data}
	\vspace{-6mm}
\end{figure*}


We generate  100 data points with 3 attributes:
2 real-valued and non-sensitive attributes $X1$ and $X2$
and 1 binary attribute $A$ which serves
as the protected attribute.
We first draw two-dimensional datapoints from a mixture of Gaussians with two components: (i) isotropic Gaussian with unit variance and (ii) correlated Gaussian with covariance 0.95 between the two attributes
and variance 1 for each attribute. 
To study the influence of membership to the protected group
(i.e., $A$ set to 1), 
we generate three variants
of this data:
\squishlist
	\item Random: $A$ is set to 1 with probability $0.3$ at random.
	\item Correlation with $X1$:
$A$ is set to $1$ if $X1 \leq 3$.
	\item Correlation with $X2$: 
$A$ is set to $1$ if $X2 \leq 3$. 
\squishend
So the three synthetic datasets have the same values for the non-sensitive attributes $X1$ and $X2$ as well for the outcome variable $Y$. The datapoints 
differ only on membership to the protected group and its distribution
across output classes $Y$. 


Figure \ref{fig:synthetic-data} shows these three cases row-wise:
subfigures a-c, d-f, g-i, respectively.
The left column of the figure displays the original data,
with the two labels for output $Y$ depicted by marker: ``o'' for $Y=0$ and ``+'' for $Y=1$
and the membership to the protected group by color: orange for $A=1$ and blue for $A=0$.
The middle column of 
Figure \ref{fig:synthetic-data} shows the learned {\em iFair}
representations, and the right column shows the representations
based on \emph{LFR}. 
Note that the values of importance in Figure \ref{fig:synthetic-data} (middle and right column) are 
the positions of the data points in the two-dimensional 
latent space and the classifier decision boundary (solid line). The color of the datapoints and the markers (o and +) depict the true class and true group membership, and not the learned values. They are 
visualized to aid the reader in relating original data with transformed representations. 
Furthermore, small differences in the learned representation are expected due to random initializations of model parameters.
The solid line in the charts denotes the predicted classifiers' decision boundary applied on the learned representations. Hyper-parameters for both {\em iFair} as well as \emph{LFR} are chosen by performing a grid search on the set $\{0, 0.05, 0.1, 1, 10, 100\}$ for optimal individual fairness of the classifier. 
For each of the nine cases, we indicate the resulting classifier accuracy $Acc$, 
individual fairness in terms of consistency $yNN$ 
with regard to the $k=10$ nearest neighbors \cite{zemel2013learning}
(formal definition given in Section \ref{section:fairness-measures}),
the statistical parity $Parity$ with regard to the protected group $A=1$,
and equality-of-opportunity $EqOpp$ \cite{hardt2016equality} notion of group fairness.

%
%

\noindent{\bf Main findings:}  Two major insights from this study are: (i)
representations learned via \emph{iFair} remain nearly the same irrespective of 
 changes in group membership, and
(ii) \emph{iFair} significantly outperforms \emph{LFR}
on accuracy, consistency and equality of opportunity,
whereas \emph{LFR} wins on statistical parity. In the following we further discuss these findings
and their implications.

\noindent{\bf Influence of Protected Group:} The middle column in Figure \ref{fig:synthetic-data} shows that the {\em iFair} representation remains largely unaffected by the changes in the
group memberships of the datapoints. 
In other words, changing the value of the protected attribute of a datapoint, all other attribute values remaining the same, has hardly any influence on its learned representation;
consequently it has nearly no influence on the outcome 
made by the decision-making algorithms trained on these representations. 
This is an important and interesting characteristic to have in a fair representation,
as it directly relates to the definition of \emph{individual fairness}.  
In contrast, the membership to the protected group has a pronounced influence on the learned representation of the \emph{LFR} model (refer to Figure \ref{fig:synthetic-data} right column). Recall that the color of the datapoints as well as the markers (o and +) are taken from the original data. They depict the true class and membership to group of the datapoints, and are visualized to aid the reader.


\noindent{\bf Tension in Objective Function:}
The optimization via \emph{LFR} \cite{zemel2013learning} has three components: 
classifier accuracy as utility metric, individual fairness in terms of data loss, and group fairness in terms of statistical parity. 
We observe that 
by pursuing group fairness and individual fairness together, the tension with utility is very pronounced. The learned 
representations are stretched on the compromise over all three goals,
ultimately leading to sacrificing utility.
In contrast, {\em iFair} pursues only utility and individual fairness, and disregards group fairness.
This helps to make the multi-objective optimization more tractable. 
{\em iFair} clearly outperforms {\em LFR} not only on accuracy, with better decision boundaries,
but also wins in terms of individual fairness.
%
This shows that the tension between utility and individual fairness is lower than between utility and
group fairness.

\noindent{\bf Trade-off between Utility and Individual Fairness:} 
The improvement that {\em iFair} achieves 
in individual fairness comes at the expense of a small drop in utility.
The trade-off is caused by 
the loss of information in learning representative prototypes. 
The choice of the mapping function in Equation \ref{eq-prototype-mapping}
and the pairwise distance function $d(.)$ in Definition 
\ref{eq:distance-function}
affects the ability to learn prototypes. 
Our framework is flexible and easily supports other kernels and distance functions.  
Exploring these influence factors is a direction for future work.

\section{Experiments}\label{section:experiments}

The key hypothesis that we test in the experimental 
evaluation is whether {\em iFair} can indeed
reconcile the two goals of {\em individual fairness} and
{\em utility} reasonably well.
As {\em iFair} is designed as an application-agnostic
representation, we test its versatility by studying
both classifier and learning-to-rank use cases,
in Subsections \ref{subsec:classification}
and \ref{subsec:ranking}, respectively.
We compare {\em iFair} to a variety of baselines
including LFR \cite{zemel2013learning} for classification and
FA*IR \cite{zehlike2017fa} for ranking.
Although group fairness and their underlying
legal and political constraints are not among the
design goals of our
approach, we include group fairness measures
in reporting on our experiments -- shedding light
into this aspect from an empirical perspective.

\subsection{Datasets} \label{section:groundtuth}

\noindent We apply the {\em iFair} framework to five real-world, publicly available datasets,
previously used in the literature on algorithmic fairness.
\begin{itemize}
\item \textbf{ProPublica's COMPAS} recidivism dataset \cite{angwin2016machine}, a widely
used test case for fairness in machine learning and algorithmic decision making. 
We set \emph{race} as a {protected attribute}, and use the binary indicator of recidivism as the outcome variable $Y$. 
\item \textbf{Census Income} dataset consists of survey results of income of 48,842 adults in the US \cite{Dua:2017}. We use gender as the protected attribute and the binary indicator variable of \emph{income $> 50 K$} as the outcome variable Y.

\item \textbf{German Credit} data has $1000$ instances of credit risk assessment records \cite{Dua:2017}. Following the literature, we set \emph{age} as the sensitive attribute, and \emph{credit worthiness} as the outcome variable.

\item \textbf{Airbnb} data consists of house listings from five major cities in the US, collected from \url{http://insideairbnb.com/get-the-data.html} (June 2018). 
After appropriate data cleaning, there are $27,597$ records.
For experiments, we choose a subset of 22 informative attributes (categorical and numerical) and infer host gender from \emph{host name}, using lists of common first names. 
We use \emph{gender} of the host as the protected attribute and \emph{rating/price} as the ranking variable.

\item \textbf{Xing} is a popular job search portal in Germany (similar to LinkedIn).
We use the anonymized data given by \cite{zehlike2017fa}, 
consisting of top $40$ profiles returned for $57$ job queries. 
For each candidate we collect information about job category, work experience, education experience, 
number of views of the person's profile, and gender. 
We set \emph{gender} as the protected attribute. We use a weighted sum of work experience, education experience and number of profile views as a score that serves as the ranking variable.
\end{itemize}

The Compas, Census and Credit datasets are used for experiments on classification,
and the Xing and Airbnb datasets are used for experiments on learning-to-rank regression. 
Table \ref{tbl:dataset-statistics} gives details of experimental settings and statistics for each dataset, including 
base-rate (fraction of samples belonging to the positive class, for both the protected group and
its complement), 
and dimensionality M (after unfolding categorical attributes). 
We choose the protected attributes and outcome variables to be in line with the literature. 
In practice, however, such decisions would be made by domain experts and according to official policies and regulations. 
The flexibility of our framework allows for multiple protected attributes, multivariate outcome variable, as well as inputs of all data types.

\begin{table}[tbh!]
	\center \setlength\tabcolsep{1.75 pt} 
	\noindent\small\resizebox{\linewidth}{!}{%
\begin{tabular}{lccccll}
	\toprule
	Dataset &  Base-rate & Base-rate & N & M & Outcome & Protected\\
     & protected & unprotected & & \\
	\midrule
	Compas &        0.52 &         0.40 & 6901 & 431  & recidivism & race\\
	Census &        0.12 &         0.31 & 48842 & 101  & income & gender\\
	Credit &        0.67 &         0.72 & 1000 & 67  & loan default & age\\
	Airbnb &        - &        - & 27597 & 33 & rating/price & gender\\
	Xing &        - &         - & 2240 & 59  & work + education  & gender\\
	\bottomrule
\end{tabular}
}
\caption{Experimental settings and statistics of the datasets.}
\label{tbl:dataset-statistics}
\vspace{-5mm}
\end{table}

\subsection{Setup and Baselines} 
\noindent In each dataset, categorical attributes are transformed using one-hot encoding, and all features vectors are normalized to have unit variance. We 
randomly split the datasets into three parts. We use one part to train the model to learn model parameters, the second part as a validation set to choose hyper-parameters by performing a grid search (details follow), and the third part as a test set. We use the same data split to compare all methods.

We evaluate all data representations -- {\em iFair} against various baselines --
by comparing the 
results of a standard classifier (\emph{logistic regression}) and 
a learning-to-rank regression model (\emph{linear regression}) applied to
\begin{itemize}
\item {\bf Full Data}: the original dataset.
\item {\bf Masked Data}: the original dataset without protected attributes.
\item {\bf SVD}: transformed data by performing dimensionality reduction via singular value decomposition (SVD) \cite{halko2009finding}, with two variants of data: 
(a) full data and (b) masked data. 
We name these variants \emph{\bf SVD} and \emph{\bf SVD-masked}, respectively.
\item \emph{\bf LFR}: the learned representation by the method of \citet{zemel2013learning}.
\item \emph{\bf FA*IR}: this baseline does not produce any data representation. FA*IR \cite{zehlike2017fa} is a ranking method which expects as input a set of candidates ranked by their \emph{deserved scores} and returns a ranked permutation which satisfies group fairness at every prefix of the ranking. We extended the code shared by \citet{zehlike2017fa} 
to make it suitable for comparison (see Section \ref{section:ranking-experiments}).
\item \emph{\bf iFair}: the representation learned by our model. We perform experiments 
with two kinds of initializations for the model parameter $\alpha$ (attribute weight vector):
(a) random initialization in $(0,1)$ and
(b) initializing protected attributes to (near-)zero values, to reflect the
intuition that protected attributes should be discounted in the distance-preservation
of individual fairness (and avoiding zero values to allow slack for the numerical computations in
learning the model).
We call these two methods \emph{\bf iFair-a} and \emph{\bf iFair-b}, respectively.
\end{itemize}

\noindent {\bf Model Parameters:} All the methods were trained in the same way. 
We initialize model parameters ($v_k$ vectors and the $\alpha$ vector)
to random values from uniform distribution in $(0,1)$
(unless specified otherwise, for the {\em iFair-b} method).
To compensate for variations caused due to initialization of model parameters, for each method
and at each setting, we report the results from the best of $3$ runs. 

\noindent{\bf Hyper-Parameters:} As for hyper-parameters (e.g., $\lambda$ and $\mu$ in Equation \ref{eq:objective} of {\em iFair}),
including the dimensionality $K$ of the low-rank representations,
we perform a grid search over the set $\{0, 0.05, 0.1, 1, 10, 100\}$ for mixture coefficients and
the set $\{10,20,30\}$ for the dimensionality $K$. 
Recall that the input data is pre-processed with categorical attributes 
unfolded into binary attributes;
hence the choices for $K$.

The mixture coefficients ($\lambda, \mu, \dots$) control the trade-off between
different objectives: utility, individual fairness, group fairness (when applicable).
Since it is all but straightforward to decide which of the multiple objectives is more important, 
we choose these hyper-parameters based on different choices for the
optimization goal (e.g., maximize utility alone or maximize a combination of
utility and individual fairness). 
Thus, our evaluation results report multiple observations for each model,
depending on the goal for tuning the hyper-parameters. 
When possible, we identify Pareto-optimal choices with respect to multiple objectives;
that is, choices that are not consistently outperformed by other choices for all
objectives.

\subsection{Evaluation Measures}\label{section:fairness-measures}
\squishlist
	\item \textbf{Utility:} measured as accuracy (Acc) and the area under the ROC curve (AUC) for the classification task, and as Kendall's Tau (KT) and mean average precision at 10 (MAP) for the learning-to-rank task. 
	\item \textbf{Individual Fairness:} measured as the \emph{consistency} of the outcome $\hat{y}_i$ of an individual with the outcomes of his/her k=10 nearest neighbors. 
This metric has been introduced by \cite{zemel2013learning}
\footnote{Our version slightly differs from that in \cite{zemel2013learning} by fixing a minor bug in the formula.}
and captures the intuition that similar
individuals should be treated similarly.
Note that nearest neighbors of an individual, $kNN(x_i)$, are computed 
on the original attribute values of $x_i$ excluding protected attributes, 
whereas the predicted response variable $\hat{y}_i$ is computed on the output of the learned
representations $\tilde{x_i}$.
	\begin{equation*}
	\text{yNN} = 1 - \frac{1}{M}\cdot\frac{1}{k} \cdot\sum\limits_{i = 1}^{M} \smashoperator \sum_{j \in kNN(x^*_i)} \norm{\hat{y}_i - {\hat{y}_j}}
	\end{equation*}
	\item \textbf{Group Fairness:} measured as 
\squishlist
\item[-] \emph{Equality of Opportunity} (\emph{EqOpp}) \cite{hardt2016equality}: 
the difference in the \emph{True Positives} rates
between the 
the protected group $X^+$ and the non-protected group  $X^-$;
\item[-] \emph{Statistical Parity} defined as: 
	\begin{equation*}
	Parity = 1 - \norm{\frac{1}{\norm{X^+}} \sum_{i \in X^+} \hat{y}_i - \frac{1}{\norm{X^-}}\sum_{j \in X^-}{\hat{y}_j}}
	\end{equation*}	
\squishend
We use the modern notion of \emph{EqOpp} as our primary metric of group fairness,
but report the traditional measure of \emph{Parity} as well.
\squishend

\subsection{Evaluation on Classification Task}
\label{subsec:classification}

\begin{figure*}[t!]
	\centering	
	\includegraphics[scale=0.45]{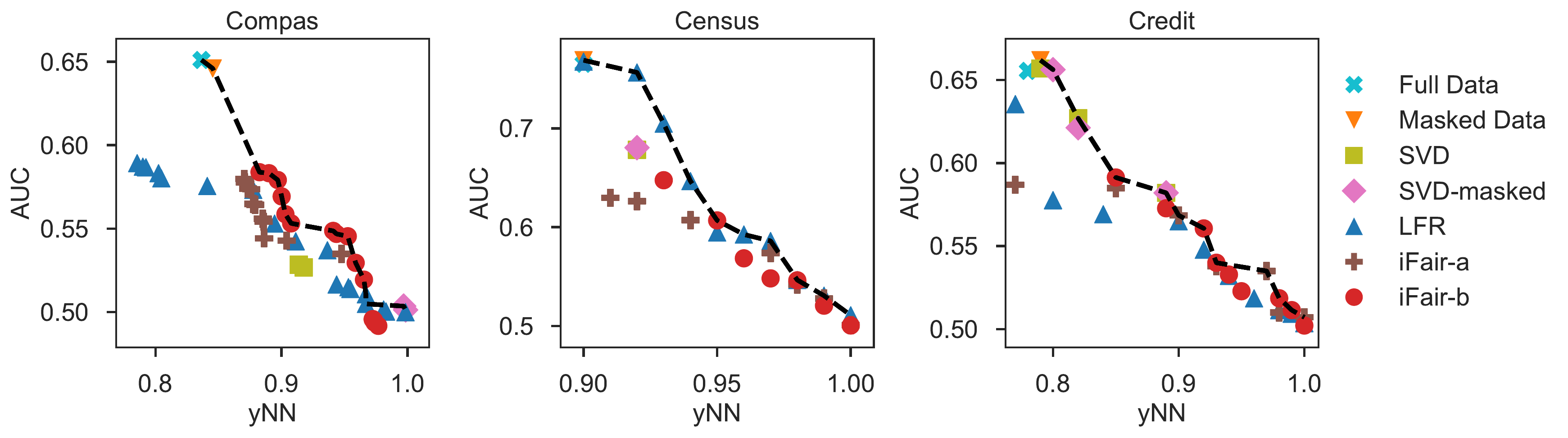}
	\caption{Plots of utility vs. individual fairness trade-off for classification task. Dashed lines represent Pareto-optimal points. 
	}
	\label{fig:compas_paretoPlot}
	\vspace{-2mm}
\end{figure*}

\begin{table*}[t!]
	\center \setlength\tabcolsep{4pt} 
	\noindent\small\resizebox{\linewidth}{!}{%
	\begin{tabular}{c|l|rrrrr|rrrrr|rrrrr}
		\toprule
		Tuning &	Method & \multicolumn{5}{c}{Compas} & \multicolumn{5}{c}{Census} & \multicolumn{5}{c}{Credit} \\
		        &           & Acc & AUC & EqOpp & Parity &  yNN &   Acc & AUC & EqOpp & Parity &  yNN &   Acc & AUC & EqOpp & Parity &  yNN \\
		\midrule
		Baseline	&	  Full Data &       0.66 & 0.65 &  0.70 &   0.72 & 0.84 &   0.84 & 0.77 &  0.90 &   0.81 & 0.90 &   0.74 & 0.66 &  0.82 &   0.81 & 0.78 \\
		\midrule
		Max			&	    LFR     &       \bf 0.60 & \bf 0.59 &  0.60 &   0.62 & 0.79 &   \bf 0.81 & \bf 0.77 &  0.81 &   0.75 & 0.90 &   0.71 & \bf 0.64 &  0.78 &   0.77 & 0.77 \\
		Utility		&	    iFair-a &       \bf 0.60 & 0.58 &  \bf 0.91 &   \bf 0.91 & 0.87 &   0.78 & 0.63 &  \bf 0.96 &   \bf 0.91 & 0.91 &   0.69 & 0.61 &  0.84 &   0.86 & 0.74 \\
		(a)			&	    iFair-b &       0.59 & 0.58 &  0.84 &   0.84 & \bf 0.88 &   0.78 & 0.65 &  0.78 &   0.85 & \bf 0.93 &   \bf 0.73 & 0.59 &  \bf 0.97 &   \bf 0.98 & \bf 0.85 \\
		\midrule
		Max			&	    LFR     &       0.54 & 0.51 &  \bf 0.99 &   0.99 & \bf 0.97 &   \bf 0.76 & 0.51 &  \bf 1.00 &   0.99 & \bf 1.00 &   0.72 & 0.51 &  \bf 0.99 &   0.98 & 0.98 \\
		Fairness	&	    iFair-a &       \bf 0.56 & \bf 0.53 &  0.97 &   0.99 & 0.95 &   \bf 0.76 & 0.51 &  0.95 &   \bf 1.00 & 0.99 &   \bf 0.73 & \bf 0.53 &  \bf 0.99 &   0.98 & 0.97 \\
		(b)			&	    iFair-b &       0.55 & 0.52 &  0.98 &   \bf 1.00 & \bf 0.97 &   \bf 0.76 & \bf 0.52 &  0.98 &   0.99 & 0.99 &   0.72 & 0.51 &  \bf 0.99 &   \bf 1.00 & \bf 0.99 \\
		\midrule
					&	    LFR     &       0.59 & 0.57 &  0.72 &   0.77 & 0.88 &   \bf 0.78 & \bf 0.76 &  \bf 0.94 &   0.74 & 0.92 &   0.71 & \bf 0.64 &  0.78 &   0.77 & 0.77 \\
		Optimal		&	    iFair-a &       \bf 0.60 & \bf 0.58 &  \bf 0.91 &   \bf 0.91 & 0.87 &   0.77 & 0.63 &  0.93 &   \bf 0.90 & 0.92 &   \bf 0.73 & 0.57 &  0.94 &   0.94 & \bf 0.90 \\
		(c)			&	    iFair-b &       0.59 & \bf 0.58 &  0.83 &   0.84 & \bf 0.89 &   \bf 0.78 & 0.65 &  0.78 &   0.85 & \bf 0.93 &   \bf 0.73 & 0.59 &  \bf 0.97 &   \bf 0.98 & 0.85 \\
		\bottomrule
	\end{tabular}

}
\caption{ Comparison of IFR vs iFair for Classification task, with hyper parameter tuning for criterion (a) max utility: best AUC (b) best Individual Fairness: best consistency, 
and (c) ``Optimal'': best harmonic mean of AUC and consistency.}
\label{tbl:classification_task}
\vspace{-3mm}
\end{table*}

This section evaluates the effectiveness of {\em iFair} and its competitors
on a classification task. We focus on the utility-(individual)fairness tradeoff that
learned representations alleviate when used to train classifiers.
For all methods, wherever applicable, hyper-parameters were tuned via grid search. Specifically, we chose the models that were Pareto-optimal with regard to AUC and yNN. 

\noindent{\bf Results:} Figure \ref{fig:compas_paretoPlot} shows the result for all methods and datasets,
plotting utility (AUC) against individual fairness (yNN). The dotted lines show models that are Pareto-optimal with regard to AUC and yNN.
We observe that there is a considerable amount of unfairness
in the original dataset, which is reflected in the results of \emph{Full Data} in Figure \ref{fig:compas_paretoPlot}. 
\emph{Masked Data} and the two SVD variants shows an improvement in fairness; however, there is 
still substantial unfairness hidden in the data in the form of correlated attributes. 
For the Compas dataset, which is the most difficult of the three datasets due to its dimensionality, SVD  completely fails.
The representations learned by  \emph{LFR} and {\em iFair} 
dominate all other methods in coping with the trade-off.
\emph{iFair-b} is the overall winner: it is consistently Pareto-optimal for all
three datasets and all but the degenerate extreme points.
For the extreme points in the trade-off spectrums, no method can achieve
near-perfect utility without substantially losing fairness and no method can be
near-perfectly fair without substantially losing utility.

Table \ref{tbl:classification_task} shows detailed results for three choices of
tuning hyper-parameters (via grid search):
(a) considering utility (AUC) only,
(b) considering individual fairness (yNN) only,
(e) using the harmonic mean of utility and individual fairness
as tuning target.
Here we focus on the \emph{LFR} and \emph{iFair} methods, as the other
baselines do not have hyper-parameters to control trade-offs
and are good only at extreme points of the objective space anyway.
The results confirm and further illustrate the findings of Figure \ref{fig:compas_paretoPlot}.
%
The two \emph{iFair} methods, tuned for the combination of utility and individual fairness
(case (c)), achieve the best overall results: iFair-b shows an improvement of 6 percent in consistency,
for a drop of 10 percent in Accuracy for Compas dataset. 
(+3.3\% and -7\% for Census, and +9\% and -1.3\% for Credit). 
Both variants of \emph{iFair} outperform \emph{LFR} by achieving significantly better individual
fairness, with on-par or better values for utility.


\subsection{Evaluation on Learning-to-Rank Task} \label{section:ranking-experiments}
\label{subsec:ranking}
This section evaluates the effectiveness of {\em iFair} 
on a regression task for ranking people on Xing and Airbnb dataset.  
We report ranking utility in terms of Kendall's Tau (KT), average precision (AP), individual fairness in terms of consistency (yNN) and group fairness in terms of fraction of protected candidates in top-10 ranks (statistical parity equivalent for ranking task).
To evaluate models in a real world setting, for each dataset we constructed multiple queries and corresponding ground truth rankings. In case of  Xing dataset we follow \citet{zehlike2017fa} and use the 57 job search queries. For Airbnb dataset, we generated a set of queries based on attributes values for \emph{city}, \emph{neighborhood} and \emph{home type}. After filtering for queries which had at least 10 listings we were left with 43 queries.

As stated in Section \ref{section:groundtuth}, for the Xing dataset,
the deserved score is a weighted sum of the true qualifications of an individual,
that is, work experience, education experience and number of profile views. 
To test the sensitivity of our results for different choices of weights, 
we varied the weights over a grid of values in $[0.0, 0.25, 0.5, 0.75, 1.0]$. 
We observe that the choice of weights has no significant effect on the measurs of interest. 
Table \ref{tbl:ad-hoc_score_test} shows details. 
For the remainder of this section, the reported results correspond to uniform weights.

\begin{table}[tbh!]
	\center \setlength\tabcolsep{4pt} 
	\noindent\small\resizebox{\linewidth}{!}{%
		\begin{tabular}{ccc |c|cccc}
			\toprule
			\multicolumn{3}{c}{Weights} &  Base-rate & MAP &  KT &  yNN  & \% Protected  \\
			$\alpha_{work}$ &    $\alpha_{edu}$ &  $\alpha_{views}$ & Protected  &   &   &  & in output\\
			\midrule
			0.00 & 0.50 & 1.00 &      33.57 & 0.76 &     0.58 &      1.00 &                     31.07 \\
			0.25 & 0.75 & 0.00 &      33.57 & 0.83 &     0.69 &      0.95 &                     35.54 \\
			0.50 & 1.00 & 0.25 &      32.68 & 0.74 &     0.56 &      1.00 &                     31.07 \\
			0.75 & 0.00 & 0.50 &      32.68 & 0.75 &     0.55 &      1.00 &                     31.07 \\
			0.75 & 0.25 & 0.00 &      31.25 & 0.84 &     0.74 &      0.96 &                     33.57 \\
			1.00 & 0.25 & 0.75 &      32.86 & 0.75 &     0.56 &      1.00 &                     31.07 \\
			1.00 & 1.00 & 1.00 &      32.68 & 0.76 &     0.57 &      1.00 &                     31.07 \\
			\bottomrule
		\end{tabular}
	}
	\caption{Experimental results on sensitivity of iFair to weights in ranking scores for Xing dataset.} 
	\label{tbl:ad-hoc_score_test}
	\vspace{-3mm}
\end{table}

Note that the baseline
\emph{LFR} used for the classification experiment, is not geared for regression tasks and thus omitted here.
Instead, we compare {\em iFair} against the \emph{FA*IR} method of
\citet{zehlike2017fa}, which is specifically designed to incorporate group fairness into rankings.

\noindent \textbf{Baseline FA*IR:} This ranking method takes as input a set of candidates ranked according to a precomputed score, and returns a ranked permutation which satisfies group fairness without making any changes to the scores of the candidates. Since one cannot measure consistency directly on rankings, we make a minor modification to FA*IR such that it also returns fair scores along with a fair ranking. 
To this end, we feed masked data to a linear regression model and compute a score for each candidate. 
FA*IR operates on two priority queues (sorted by previously computed scores): $P_0$ for non-protected candidates and $P_1$ for protected candidates. For each rank k, it computes the minimum number of protected candidates required to satisfy statistical parity (via significance tests) at position k.
If the parity constraint is satisfied, it chooses the best candidate and its score from $P_0$ $\cup$ $P_1$. 
If the constraint is not satisfied, it chooses the best candidate from P1 for the next rank and leaves a placeholder for the score. 
Our extension linearly interpolates the scores to fill the placeholders, 
and thus returns a ranked list along with ``fair scores''. 

\noindent{\bf Results:} Table \ref{tbl:Xing_comparison} shows a comparison of experimental results for 
the ranking task for all methods across all datasets. We report mean values of average precision (MAP),
Kendall's Tau (KT) and consistency (yNN) over all 57 job search queries for Xing and 43 house listing queries for Airbnb.
Similar to the classification task, \emph{Full Data} and \emph{Masked Data} have the best utility (MAP and KT), whereas iFair has the best individual fairness (yNN). iFair clearly outperforms both variants of SVD by achieving significantly better individual fairness (yNN) for 
comparable values of utility. As expected, \emph{FA*IR}, which optimizes to satisfy statistical parity across groups, has the highest fraction of protected candidates in the top 10 ranks, but does not achieve
any gains on individual fairness. This is not surprising, though, given its design goals.
It also underlines our strategic point
that individual fairness needs to be
explicitly taken care of as a first-order objective.
Between {\em FA*IR} and {\em iFair}, there is no 
clear winner,
given their different objectives.
We note, though, that the good utility that {\em FA*IR} achieves in some
configurations critically hinges on the choice of the value for its parameter $p$.

\begin{table}[tbh!]
	\center \setlength\tabcolsep{2pt} 
	\noindent\tiny\resizebox{\columnwidth}{!}{%
\begin{tabular}{clcccc}
	\toprule
	Dataset &             Method &   MAP &  KT &  yNN & \% Protected\\
	 &              &  (AP@10)  &  (mean) &  (mean) &  in top $10$ \\
	\midrule
 &      Full Data & \textbf{1.00} &     \textbf{1.00} &      0.93 &                     32.50 \\
 &    Masked Data & \textbf{1.00} &     \textbf{1.00} &      0.93 &                     32.68 \\
 &            SVD & 0.74 &     0.59 &      0.81 &                     31.79 \\
Xing &     SVD-masked & 0.67 &     0.50 &      0.78 &                     32.86 \\
 (57 queries)&          FA*IR (p = 0.5) & 0.93 &     0.94 &      0.92 &                    38.21 \\
 					 & FA*IR (p = 0.9)          & 0.78          & 0.78          & 0.85       &  \textbf{48.57}   \\
 &        iFair-b & 0.76 &     0.57 &      \textbf{1.00} &                     31.07 \\
\midrule
	 &          Full Data & \textbf{0.68} &     \textbf{0.53} &      0.72 &                     47.44 \\
	 &        Masked Data & 0.67 &     0.53 &      0.72 &                     47.44 \\
	 &                SVD & 0.66 &     0.49 &      0.73 &                     48.37 \\
	 Airbnb &         SVD-masked & 0.66 &     0.49 &      0.73 &                     48.37 \\	
	 (43 queries) &  FA*IR (p = 0.5) & 0.67 &     0.52 &      0.72 &                     48.60 \\
                 &  FA*IR (p = 0.6) & 0.65 &     0.51 &      0.73 &                     \textbf{51.16} \\
	 &            iFair-b & 0.60 &     0.45 &     \textbf{0.80} &                     49.07 \\
\bottomrule
\end{tabular}
}
	\caption{Experimental results for ranking task. Reported values are means over multiple query rankings for the criterion ``Optimal'': best harmonic mean of MAP and yNN.} 
	\label{tbl:Xing_comparison}
	\vspace{-3mm}
\end{table}


\subsection{Information Obfuscation 
\& Relation to Group Fairness}
\begin{figure}[tbh!]
	\centering	
	\includegraphics[scale=0.48]{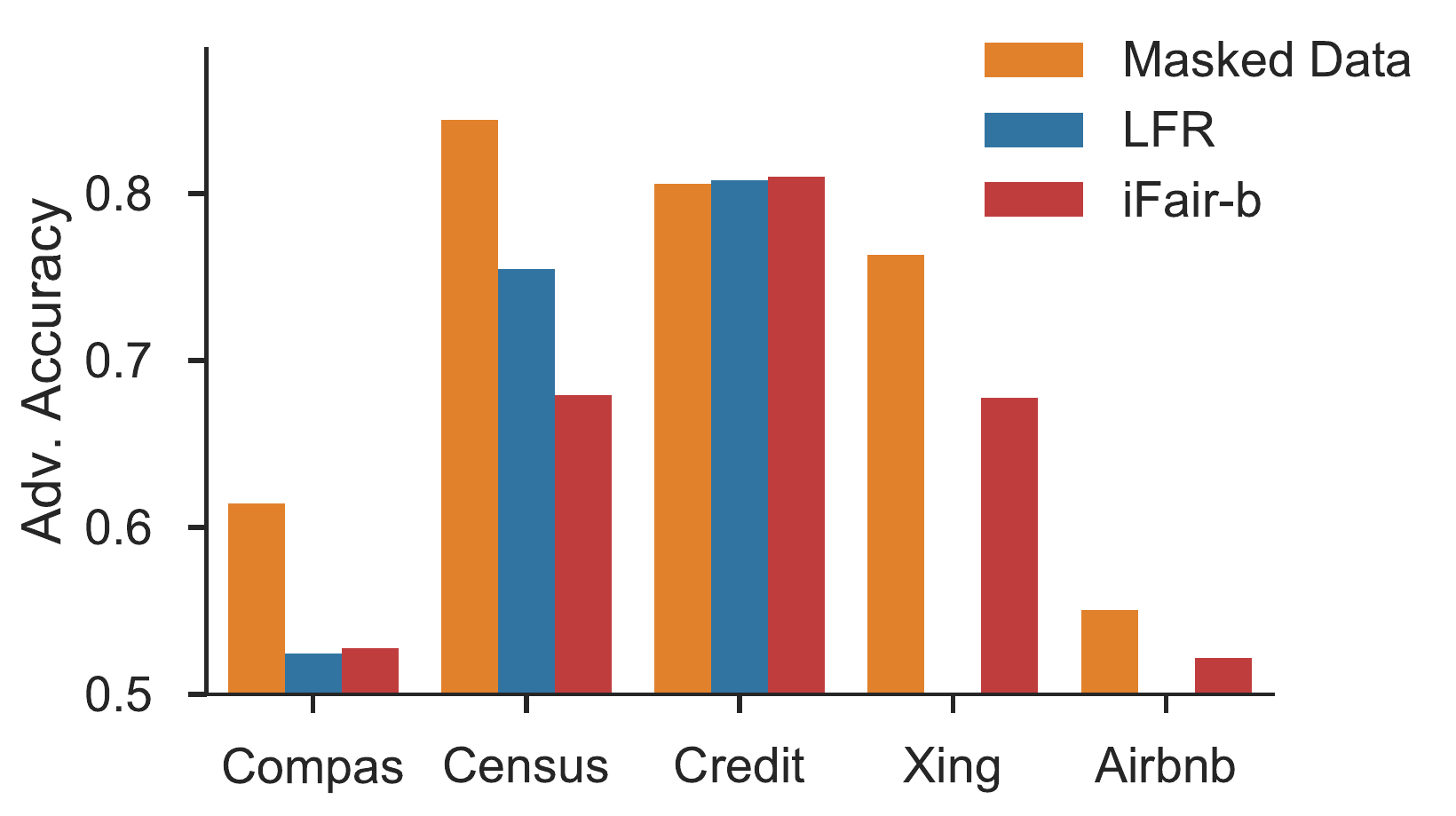}
	\vspace{-1mm}
	\caption{Adversarial accuracy of predicting protected group membership. The lower the better.}
	\label{fig:adversary}
	\vspace{-2mm}
\end{figure}

\begin{figure}[tbh!]
	\begin{subfigure}{0.98\columnwidth}
		\centering	
		\includegraphics[scale=0.46]{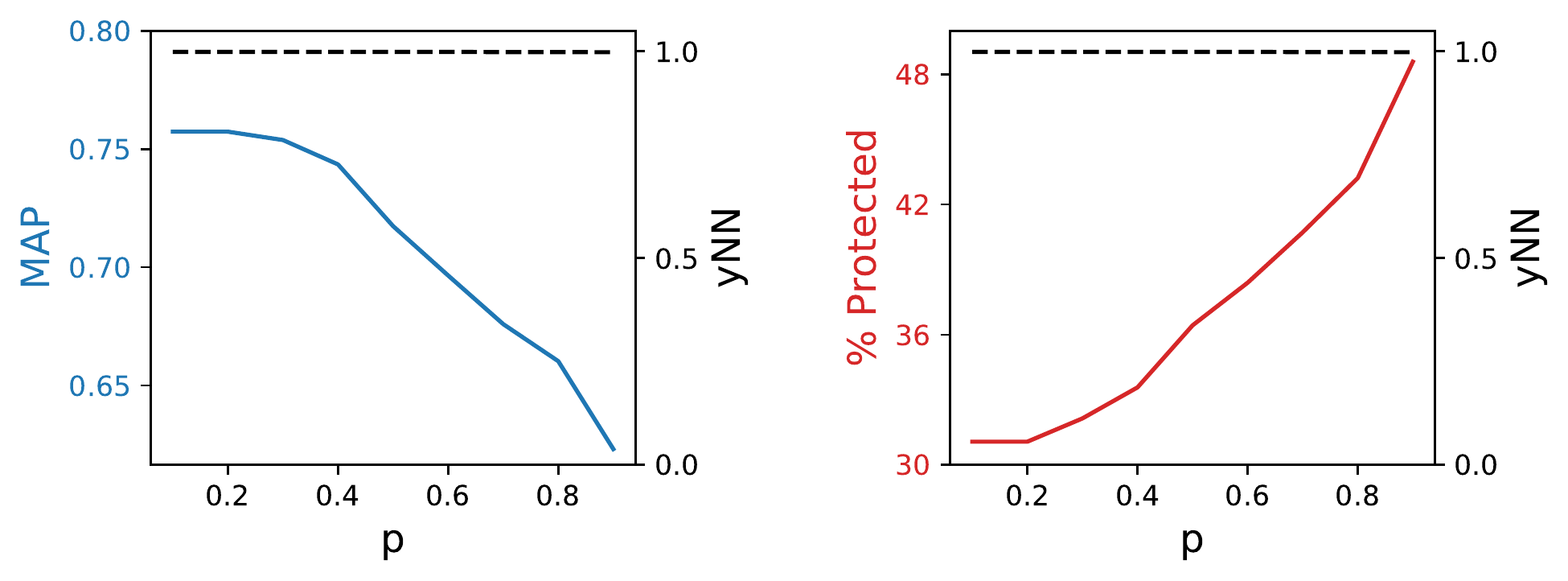}
		\vspace{-9mm}
		\subcaption{Xing}
		\label{fig:IFR_FAIR_Xing}
		\vspace{1mm}
	\end{subfigure}	
	\newline
	\begin{subfigure}{0.98\columnwidth}
		\centering	
		\includegraphics[scale=0.46]{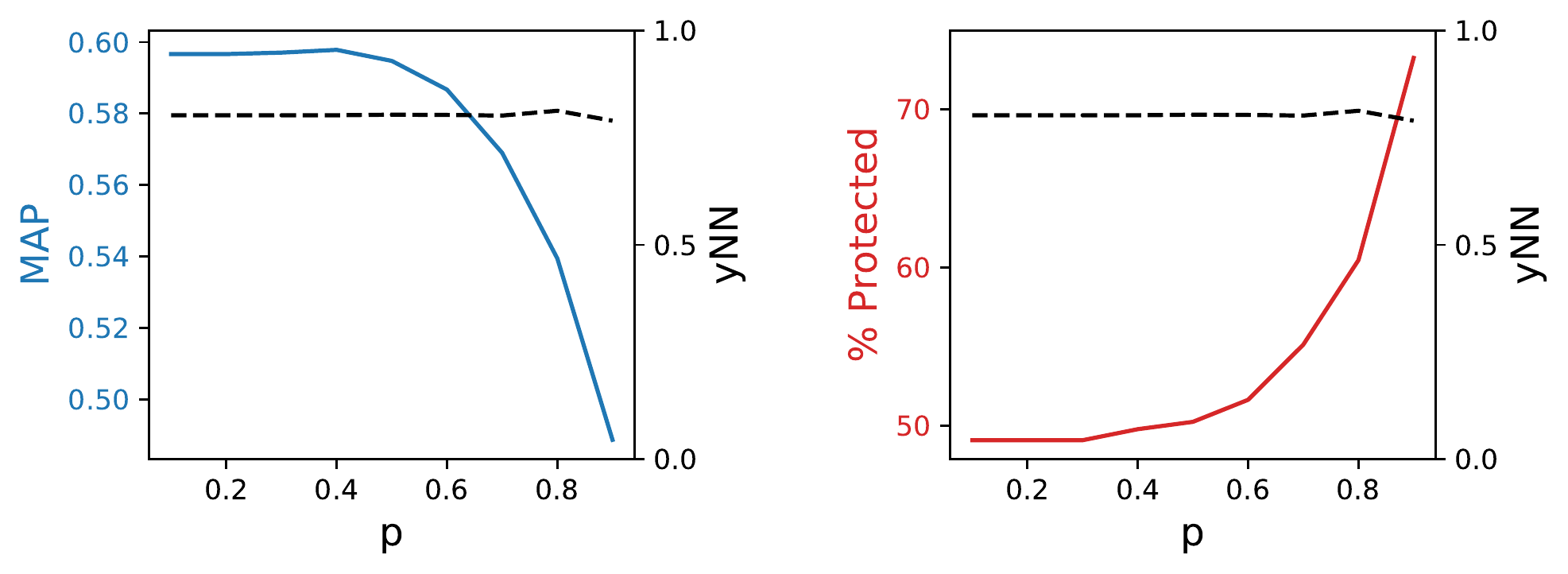}
		\vspace{-9mm}
		\subcaption{Airbnb}
		\label{fig:IFR_FAIR_Airbnb}
		\vspace{1mm}
	\end{subfigure}
	\caption{Applying FA*IR algorithm to \emph{iFair} representations.}
	\label{fig:IFR_FAIR}
	\vspace{-5mm}
\end{figure}
We also investigate the ability of our model to 
obfuscate
information about protected attributes. 
A reasonable proxy to measure the 
extent to which protected information 
is still
retained in the {\em iFair} representations
 is to predict the value of the protected attribute from the learned representations. 
We trained a logistic-regression classifier to predict the protected group membership from: (i) Masked Data (ii) learned representations via \emph{LFR}, and (iii) learned representations via \emph{iFair-b}. 
 
\spara{Results:} Figure \ref{fig:adversary} 
shows the adversarial accuracy of predicting the protected group membership for all 5 datasets (with LFR not applicable
to Xing and Airbnb). 
For all datasets, \emph{iFair} manages to 
substantially reduce the adversarial accuracy.
This signifies that its learned representations contain little information on
 protected attributes,
despite the presence of correlated attributes.
In contrast,  \emph{Masked Data} still reveals 
enough implicit information on protected groups
and cannot prevent 
the adversarial classifier from achieving fairly good accuracy.

\spara{Relation to Group Fairness:}
Consider the notions of group fairness defined in Section \ref{section:fairness-measures}. 
Statistical parity
requires the probability of predicting positive outcome to be independent of the protected attribute:
$P(\hat{Y} = 1| S = 1) = P(\hat{Y} = 1| S = 0)$. 
Equality of opportunity requires this probability to be
independent of the protected attribute conditioned on the true outcome $Y$: $P(\hat{Y} = 1| S = 1, Y = 1) = P(\hat{Y} = 1| S = 0, Y = 1)$. 
Thus, forgetting information about the protected attribute indirectly helps improving group fairness;
as algorithms trained on the 
individually fair representations carry largely reduced information
on protected attributes. 
This argument is supported by our empirical results on group fairness for all datasets. In Table \ref{tbl:classification_task}, 
although group fairness is not an explicit goal, we observe substantial improvements by 
more than 10 percentage points;
the performance for other datasets is similar.

However, the extent to which {\em iFair} also benefits
group fairness criteria 
depends on the base rates $P(Y = 1| S = 1)$ and $P(Y = 1| S = 0)$ of the underlying data. Therefore, in applications where statistical parity is a legal requirement, additional steps are needed,
as discussed next.

\spara{Enforcing parity:}
By its application-agnostic design,
it is fairly straightforward to enhance {\em iFair} by
post-processing steps to enforce 
statistical parity, if needed.
Obviously, this requires access to the values of
protected attributes, but this is the case for
all group fairness methods.

We demonstrate the extensibility of our framework 
by applying the \emph{FA*IR} \cite{zehlike2017fa} 
technique as a post-processing step to the {\em iFair} representations of the Xing and Airbnb data. 
For each dataset, we generate top-k rankings by varying the target minimum fraction of protected candidates (parameter $p$ of the FA*IR algorithm).
Figure \ref{fig:IFR_FAIR} reports ranking utility (MAP), percentage of protected candidates in top 10 positions, 
and individual fairness
(yNN) 
for increasing values of the FA*IR parameter $p$.
The key observation is that the combined model
{\em iFair + FA*IR} can indeed achieve whatever
the required share of protected group members is,
in addition to the individual fairness property of the learned representation.

\balance
\section{Conclusions}\label{section:discussion}
We propose \emph{iFair},
a generic and versatile, unsupervised framework to perform a 
probabilistic transformation of data into individually fair representations. 
Our approach accomodates 
two important criteria. 
First, we view fairness from an application-agnostic view, 
which allows us to incorporate it in a wide variety of tasks,
including general classifiers and regression for learning-to-rank. 
%
Second, we treat individual fairness as a property of the dataset (in some sense, like privacy), which can be achieved by pre-processing the data into a transformed representation.
This stage
does not need access to protected attributes. 
If desired, we can also post-process the learned representations and enforce
group fairness criteria such as statistical parity.

We applied our model to five real-world datasets, empirically
demonstrating that utility and individual fairness can be reconciled
to a large degree.
Applying classifiers and regression models to \emph{iFair} representations 
leads to 
algorithmic decisions that are substantially more consistent
than the decisions made on the original data. 
Our approach is the first method to compute individually fair results in learning-to-rank tasks. For classification tasks, it outperforms the state-of-the-art prior work.




%
%
%

\section{Acknowledgment} This research was supported by the ERC Synergy Grant ``imPACT'' (No. 610150) and ERC Advanced Grant ``Foundations for Fair Social Computing'' (No. 789373).

\bibliography{references}
\bibliographystyle{IEEEtranN}
\end{document}